\documentclass{bmvc2k}

\title{\ourmethod{}: A Convolutional Neural Network for Tetrahedral Mesh Generation}

\addauthor{William Gao}{wmg@uchicago.edu}{1}
\addauthor{April Wang}{aprilyanli38@gmail.com}{1}
\addauthor{Gal Metzer}{gal.metzer@gmail.com}{2}
\addauthor{Raymond A. Yeh}{rayyeh@purdue.edu}{3}
\addauthor{Rana Hanocka}{ranahanocka@uchicago.edu}{1}

\addinstitution{
 University of Chicago\\
 Chicago, IL, USA
}
\addinstitution{
 Tel Aviv University\\
 Tel Aviv-Yafo, Israel
}
\addinstitution{
 Purdue University\\
 West Lafayette, IN, USA
}

\runninghead{W. Gao, A. Wang, G. Metzer, R. Yeh, R. Hanocka}{TetGAN}

\usepackage{booktabs} %
\usepackage{graphicx}
\usepackage{amsmath}
\usepackage{booktabs}
\usepackage{xcolor}
\usepackage{soul}
\usepackage{wrapfig}
\usepackage{overpic}

\newcommand{\ourmethod}{TetGAN}
\newcommand{\norm}[1]{\left| \left| #1 \right| \right|}

\graphicspath{{figures/}} %

\usepackage{amsmath,amsfonts,bm}

\def\1{\bm{1}}

\def\mD{{\bm{D}}}

\def\mM{{\bm{M}}}

\def\mO{{\bm{O}}}

\def\mT{{\bm{T}}}

\def\mW{{\bm{W}}}
\def\mX{{\bm{X}}}

\def\mZ{{\bm{Z}}}

\DeclareMathAlphabet{\mathsfit}{\encodingdefault}{\sfdefault}{m}{sl}
\SetMathAlphabet{\mathsfit}{bold}{\encodingdefault}{\sfdefault}{bx}{n}
\newcommand{\tens}[1]{\bm{\mathsfit{#1}}}

\def\tT{{\tens{T}}}

\def\gI{{\mathcal{I}}}

\def\gN{{\mathcal{N}}}

\def\gS{{\mathcal{S}}}

\def\sR{{\mathbb{R}}}

\DeclareMathOperator*{\argmin}{arg\,min}

\usepackage{symbols}
\usepackage{multirow}
\usepackage{placeins}
\usepackage{tabularx}

\begin{document}

\maketitle
\begin{minipage}{0.9\linewidth}
    \small
    \centering
    \newcommand{\traincolor}{\color[rgb]{0.26,0.64,0.17}}
    \newcommand{\testcolor}{\color[rgb]{0.15,0.53,0.87}}
    \centering
    \includegraphics[width=\textwidth]{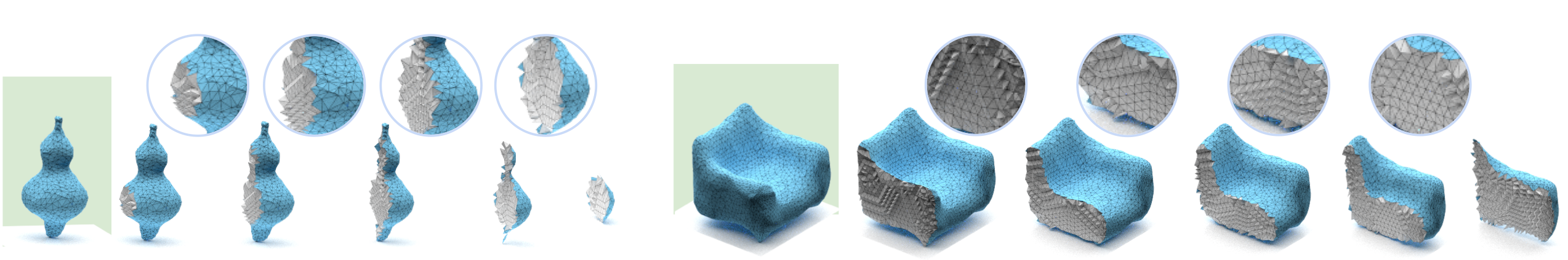}
    \vspace{-.7cm}
    \captionof{figure}{\ourmethod{} {\traincolor \textbf{synthesizes novel tetrahedral meshes}} with solid interiors.}
    \label{fig:teaser}
\end{minipage}

\begin{abstract}
We present TetGAN, a convolutional neural network designed to generate tetrahedral meshes. We represent shapes using an irregular tetrahedral grid which encodes an occupancy and displacement field. Our formulation enables defining tetrahedral convolution, pooling, and upsampling operations to synthesize explicit mesh connectivity with variable topological genus. The proposed neural network layers learn deep features over each tetrahedron and learn to extract patterns within spatial regions across multiple scales. We illustrate the capabilities of our technique to encode tetrahedral meshes into a semantically meaningful latent-space which can be used for shape editing and synthesis. Our project page is at \url{https://threedle.github.io/tetGAN/}.
\end{abstract}

\section{Introduction}
\label{sec:intro}

We approach learning to synthesize and edit 3D shapes through a neural framework designed for \textit{tetrahedral} meshes. Tetrahedral meshes are used widely across a myriad of applications, such as articulated shape deformations~\cite{wang2015linear, jacobson2011bounded}, solid modeling~\cite{cutler2002procedural}, scientific computing~\cite{uesu2005simplification}, and physics-based simulations~\cite{ftetwild, jatavallabhula2021gradsim}. However, deep learning on tetrahedral meshes is only just starting to develop~\cite{deftet, shen2021dmtet}.

We propose a framework for generating tetrahedral meshes, called \ourmethod{}. Our system learns to produce \textit{novel} and diverse tetrahedral meshes from random noise. Simultaneously, \ourmethod{} provides \textit{controlled} manipulation of existing tetrahedral meshes through latent-encoding capabilities. Our framework trains an encoder to map tetrahedral meshes into a semantically meaningful latent space. This enables performing latent space interpolation or arithmetic operations for effective 3D shape manipulations.

\begin{wrapfigure}{r}{0.12\textwidth}
\raggedleft
\vspace{-7pt}
\hspace*{-15pt}
\includegraphics[width=0.12\textwidth]{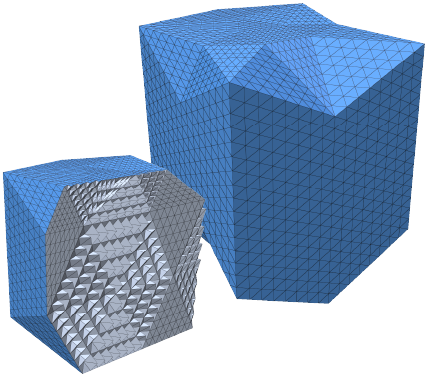}
\vspace{-10pt}
\hspace*{5pt} %
\end{wrapfigure}
\ourmethod{} leverages an irregular grid of tightly packed tetrahedra within a bounding cuboid to create a spatial structure for learning local correlations in 3D. Akin to a pixel in an image, each tetrahedron is the basic building block for learning  over the network input. To represent a 3D shape, we encode each tetrahedron with an \textit{occupancy} value and 3-dimensional \textit{displacement} vector to the closest point on the shape surface. This enables extracting a tetrahedral mesh by removing unoccupied tetrahedra, and displacing vertices on triangles on the surface \textit{boundary}.
This representation combines the advantages of implicit representations to handle varying topology, and explicit meshes to obtain a high level of detail efficiently. 

The irregular tetrahdral grid has recently started to gain traction as an alternative neural representation for 3D geometry \cite{munkberg2021nvdiffrec}, pioneered by DefTet and DMTet~\cite{deftet, shen2021dmtet}.
Using the irregular tetrahedral grid, topological holes can be inserted anywhere easily by removing unoccupied tetrahedra in the grid. Tetrahedral meshes are able to \textit{deform} each triangle by individually displacing vertices since deformed triangles remain planar \textit{and} connected to adjacent faces. Deforming the boundary vertices to coincide with the learned underlying shape will result in a smooth surface with varied topology. This seemingly simple property is non-trivial to obtain with 3D cubic voxel grids (\ie a hexahedral mesh). Displacing 4-points in a quadrilateral does not necessarily preserve planarity between adjacent quads.

We learn to represent local spatial relationships in 3D shapes (such as corners, curves, and planes), through carefully designed neural network layers. We use a \textit{tetrahedral convolution} layer which shares weights across the entire spatial field, in order to encode local shape cues. Beyond local regions, we progressively incorporate additional 3D \textit{context} through \textit{tetrahedral pooling}, reducing the spatial resolution of the input. The original spatial resolution is restored through \textit{upsampling} layers. The \ourmethod{} layers contain an inductive bias which leverages and learns from features across multiple spatial resolutions.

Our \ourmethod{} framework supports and benefits from both \emph{latent encoding} capabilities during inference, and unsupervised \textit{adversarial} losses during training. We train our model using distinct VAE and GAN objectives, which is different than typical use of VAE with GAN~\cite{vqgan, cycleGAN}.
In this work, the VAE training objective aims for tetrahedral meshes to be encoded and then exactly decoded using a reconstruction loss. 
Then, during inference the encoder %
maps a given input shape into the learned latent space, enabling performing latent-based editing operations. 
The GAN training objective aims to generate new tetrahedral meshes from noise that are not part of the training dataset. 
We achieve this by sampling a random latent code from a unit Gaussian distribution, used as input to the same decoder to produce a tetrahedral mesh using an adversarial loss (without any reconstruction loss).

\section{Related Work}
\label{sec:related}
Initially, convolutional neural networks built for mesh surfaces were focused on shape analysis tasks such as classification and segmentation~\cite{hanocka2019meshcnn, pdmeshnet, lahav2020meshwalker, subdivnet}. Later, techniques proposed synthesizing vertex locations over a surface mesh with fixed topology~\cite{pixel2mesh, liu2020neural, hertz2020deep, Hanocka2020p2m, gao2019sdm}. 
Though meshes can efficiently represent complex geometries, generating meshes with varied genus is challenging~\cite{sap}. Only a handful of works directly generate or modify surface mesh connectivity~\cite{nash2020polygen, rakotosaona2021differentiable, sharp2020pointtrinet}. Other techniques combine fixed genus mesh parts to form a more complex topology, such as a composition of primitives~\cite{tulsiani2017learning,chen2020bsp,paschalidou2021neural}, parametric patches~\cite{groueix2018papier}, or assembly of parts~\cite{yin2020coalesce}. Cubic voxel-based techniques can generate shapes with varied genus~\cite{chen2021decor}, however planar-face hexahedra cannot be displaced freely.

\textbf{Implicit Representations.}
Shapes can also be synthesized using implicit representations, which predict a signed distance or an occupancy value per input point~\cite{park2019deepsdf, chen2019learning, mescheder2019occupancy, 3dgan}. Implicit representations can generate shapes with any topological genus since the explicit mesh surface is extracted in post-process. ShapeGAN~\cite{kleineberg2020adversarial} employs DeepSDF~\cite{park2019deepsdf} as a generator to model a signed distance function, and a discriminator which evaluates the signed distance field produced by a batch of samples alongside latent information. The generated SDF values from the generator are arranged into a voxel volume or input to PointNet.

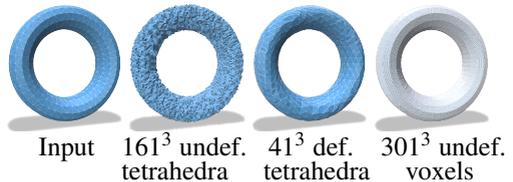
\begin{wrapfigure}{R}{0.5\textwidth}
    \vspace{-0.15cm}
    \centering
    \newcommand{\pl}{-5}
    \newcommand{\plb}{-10}
    \begin{overpic}[width=0.5\columnwidth]{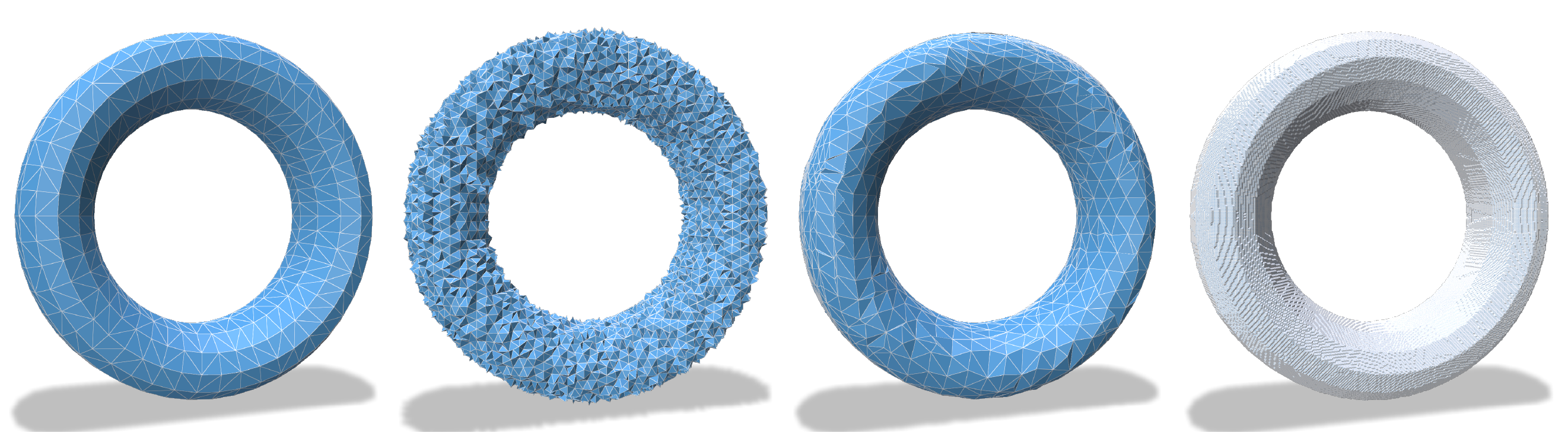}
    \put(7,  \pl){\textcolor{black}{Input}}
    \put(24,  \pl){\textcolor{black}{$161^3$ undef.}}
    \put(24,  \plb){\textcolor{black}{tetrahedra}}
    \put(54,  \pl){\textcolor{black}{$41^3$ def.}}
    \put(53,  \plb){\textcolor{black}{tetrahedra}}
    \put(77,  \pl){\textcolor{black}{$301^3$ undef.}}
    \put(82,  \plb){\textcolor{black}{voxels}}
    \end{overpic}
    \vspace{-0.05cm}
    \caption{Input torus represented with different volumetric representations. Left to right: i)  occupancies extracted from a $161^3$ tetrahedralized cube (undeformed), ii) a $~41^3$ deformable tetrahedral grid, and iii)  voxelized torus at $301^3$ extracted with marching cubes. %
    }
    \label{fig:resolution}
    \vspace{-0.5cm}
\end{wrapfigure}
\textbf{Tetrahedral meshes} are widely used in Finite Element Method (FEM) simulations~\cite{ftetwild, PRPSL15}. Computing realistic and high-quality deformations of articulated shapes commonly relies on a tetrahedral mesh~\cite{wang2015linear, jacobson2011bounded}. We use the tool QuarTet \cite{quartet} to generate the irregular grid of tetrahedra which we use to define the \ourmethod{} convolution, pooling, and upsampling layers. 
Our use of a tetrahedral grid draws inspiration from DefTet~\cite{deftet}, which learned to reconstruct tetrahedral meshes from point clouds or 2D images. Recently, Deformable Marching Tets (DMTet)~\cite{shen2021dmtet}, learns to reconstruct high-resolution surface meshes from a coarse voxel input.

One of the key differences between our tetrahedral grid representation and that of DefTet~\cite{deftet} and DMTet~\cite{shen2021dmtet, munkberg2021nvdiffrec} is how we predict and utilize displacements. In this work, our tetrahedral grid contains a vector-valued displacement to the underlying surface at each tetrahedron's centroid, which we refer to as a \emph{deformation field}. Deformation fields are a rich representation of the explicit underlying geometry, which provides shape cues over the entire grid. 
The objective of encoding and predicting a deformation field even aids the network in learning to predict more accurate occupancy fields, suggesting that deformation fields are an effective neural shape representation.
After training is complete, we leverage the network-predicted deformation field to \textit{filter} incorrect occupancy predictions %
and infer weights for \textit{smoothing} the predicted surface geometry.

While prior works apply the tetrahedral grid representation to the tasks of tetrahedral \textit{reconstruction} (DefTet) and mesh super-resolution (DMTet), our method provides \textit{generative} capabilities, a fundamentally different objective. TetGAN enables sampling novel shapes from noise, latent space interpolations, and shape editing, none of which are provided by DefTet/DMTet (nor by inverse rendering~\cite{munkberg2021nvdiffrec}). TetGAN achieves this using a novel CNN for tetrahedral meshes, with convolution/pooling blocks that are distinct from DefTet/DMTet components and tailored for the task of generation (inspired by 2D CNNs~\cite{DCGAN, stylegan, stylegan3}).
Our pooling and upsampling leverages a tetrahedral subdivision
structure to learn features over various receptive fields.
In a similar spirit, SubDivNet~\cite{subdivnet} learns using a loop subdivision structure for performing analysis tasks (\eg classification) over \textit{surface} meshes.

\textbf{Contrast to Voxels.}
Works that synthesize shapes using implicits may opt to predict over a voxel grid \cite{mescheder2019occupancy, kleineberg2020adversarial}. However, the benefits afforded by our deformation field are not trivially amenable to 3D cubic (hexahedral) voxels.
Freely displacing the four vertices per-face
\begin{wrapfigure}{r}{0.22\textwidth}
\raggedleft
\vspace{2pt}
\hspace*{-15pt}
\includegraphics[width=0.22\textwidth]{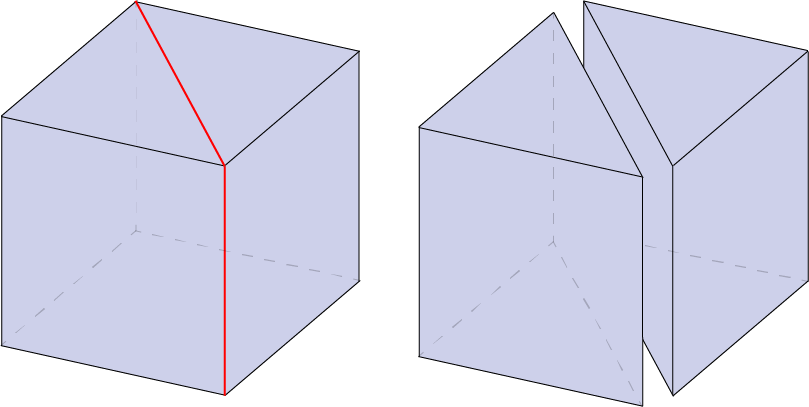}
\vspace{-16pt}
\hspace*{5pt} %
\end{wrapfigure}
will break the planar-face property.
Alternatively, fitting higher-order curved patches to 4-point sets does not guarantee connectivity between adjacent cells. Moreover, cubic cells do not trivially divide into tetrahedra, and volumetric meshing is an open active area of research~\cite{tetwild, ftetwild, TetGen}. Due to the inherent challenges in converting to tetrahedral meshes (\eg inevitable trade-offs between geometric accuracy and robustness), several works propose bypassing tetrahedral meshing altogether for a particular application~\cite{sawhney2020mcgp, Trusty_Chen_Levin_SEM_2021}.
Compared to cubic voxels, tetrahedral meshes are able to more efficiently adapt to complex geometries, such as high curvature regions as shown in Fig.~\ref{fig:resolution}. Representing a torus using different volumetric representations (with error under a threshold), requires $301^3$ undeformed cubic voxels, compared to $161^3$ undeformed tetrahedra or $41^3$ deformed tetrahedra.

\begin{figure*}[t]
    \vspace{-0.3cm}
    \centering
     \newcommand{\enccolor}{\color[rgb]{0,0,1}}
     \newcommand{\deccolor}{\color[rgb]{0.19,0.71,0.29}}
     \newcommand{\discolor}{\color[rgb]{0.98,0.69,0.23}}
    \includegraphics[width=\textwidth]{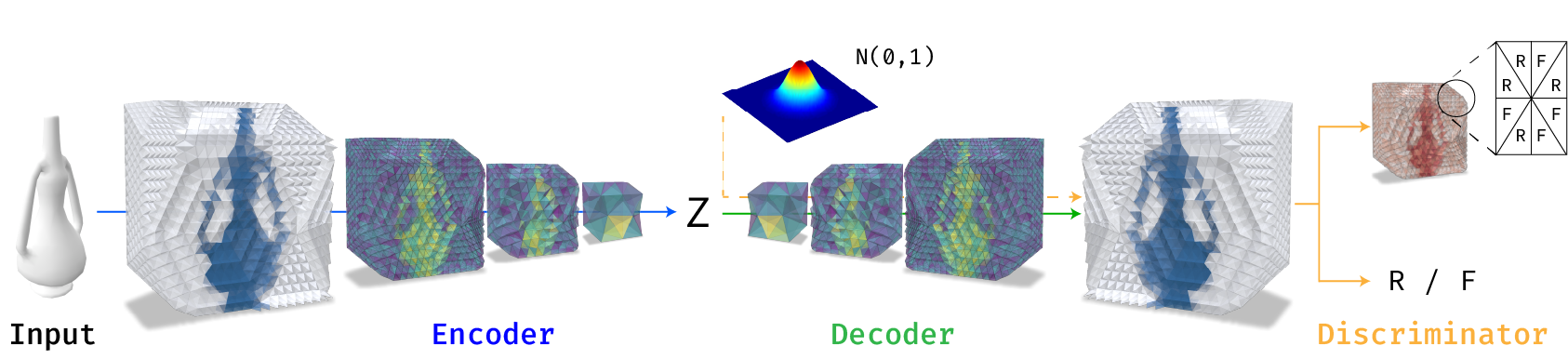}
    \small
    \vspace{-0.65cm}
    \caption{\ourmethod{} overview. We represent the input mesh using a deformation field over a tetrahedral grid. The {\enccolor \textbf{encoder}} produces a latent vector $\mZ$, which is used as input to a {\deccolor \textbf{decoder}} which performs a series of convolutions and upsampling operations. %
    The two {\discolor \textbf{discriminators}} learn to classify whether the input shape is fake (generated) or real at the patch/global level.
    } 
    \label{fig:overview}
    \vspace{-0.55cm}
\end{figure*}

\section{Method}
\label{sec:method}
We design a generative model over a packed tetrahedral grid to define a coarse inside / outside representation of a shape (\ie, an occupancy). This grid directly represents a tetrahedral mesh with no additional processing and allows an isosurface to be trivially extracted. We also define a displacement per-vertex in the grid to the underlying shape surface, which enables smoothing the coarse representation to obtain a higher fidelity result.

To learn over this representation, we propose \ourmethod{}, a deep architecture which leverages and fuses ideas from both generative adversarial neural networks (GAN)~\cite{goodfellow2014generative} and Variational Autoencoders (VAE)~\cite{vae}.
First, an encoder performs a series of convolution and pooling layers (on a tetrahedral grid) to aggregate the spatial information over the 3D shape into a latent vector. Next, a decoder processes the latent vector through a series of convolution and upsampling layers to reconstruct the 3D shape. This stage is supervised by ground truth features computed over the training data. Finally, to improve the diversity and smoothness of the learned latent space, we make use of an additional adversarial training, where the VAE decoder acts as the generator of a GAN. This enables \ourmethod{} to directly learn more point samples inside the latent space. To synthesize a new tetrahedral mesh, we sample from the learned latent space and use the decoder to generate the occupancy and displacement fields. The overall model is illustrated in~\figref{fig:overview}.

\subsection{Tetrahedral grid framework}\label{sec:tetgrid}
\textbf{Tetrahedral grid.} 
Our framework represents shapes using an occupancy and a displacement encoded within an irregular grid of tetrahedra. First, we compute an initial tetrahedralization of a cube using QuarTet~\cite{quartet} to obtain a tetrahedral grid $\tT = \{\mT_1,\hdots \mT_K\}$ consisting of $K$ tetrahedra. 
Each tetrahedron $\mT_k$ is represented by its four vertices: $\{v_{k_a}, v_{k_b}, v_{k_c}, v_{k_d}\}$, where each $v_{k_i} \in \sR^3$. To represent an input surface mesh $\mM$, we first normalize it to fit inside the unit cube. 
We calculate the occupancy value (inside or outside) at each tetrahedron's centroid. We also calculate the displacement from the tetrahedron centroid to the closest point on the mesh surface, \ie the displacement field.

\textbf{Tetrahedral convolution.} Despite being an irregular grid $\tT$, a unique structure naturally admits learning a convolution. Every tetrahedron in the grid is adjacent to exactly four tetrahedra, which defines the spatial support of the convolutional filter. Specifically, for each tetrahedron $\mT_k\in \tT$, we perform a weighted sum of $\mT_k$ and its four adjacent neighbors $\mT_j \in \tT$ (\ie $\mT_k$ and $\mT_j$ share a triangle face). A convolution on the tetrahedral grid is defined as
$\mW_0 \cdot \phi(\mT_k) + \sum_{j \in \gN_k} \mW_j \cdot \phi(\mT_j)$,
where $\gN_k$ denotes the set of adjacent neighbors of $\mT_k$, $\mW$ denotes the convolution weights, and $\phi(\mT_k)$ denotes the features for tetrahedron $\mT_k$. The ordering of the four neighbors in the convolution is prescribed by the tetrahedral grid.

\textbf{Subdivision structure for pooling and upsampling.} 
To perform pooling and upsampling on the tetrahedral grid, we need to first define how to down/upsample, for which we introduce a subdivision structure. For any $\mT_j \in \tT$, we can subdivide it into eight tetrahedra, represented as a \textit{supercell} $\gS_j=\{\mT_k\}$. A supercell is created by adding vertices at the midpoints of each edge. Starting with a low resolution grid $\tT^{(1)}$, we produce successive grids $\tT^{(2)}, \tT^{(3)}, \hdots, \tT^{(N)}$ by subdividing each tetrahedron. Naturally, pooling of features $\phi$ from $\tT^{(n)}$ to $\tT^{(n-1)}$ is defined as
{\small
\bea
\phi^{(n-1)}(\mT_j^{(n-1)}) = \mathbf{F} \left(\left\{ \phi^{(n)}(\mT_k^{(n)})\;~~\forall~ \mT_k^{(n)} \in \gS_j^{(n-1)}%
\right\}\right),
\eea
}
where $\mathbf{F}$ is a symmetric aggregation function (either a max or average).

\begin{figure}
    \begin{minipage}[c]{0.46\columnwidth}
    \centering
    \includegraphics[width=0.8\columnwidth]{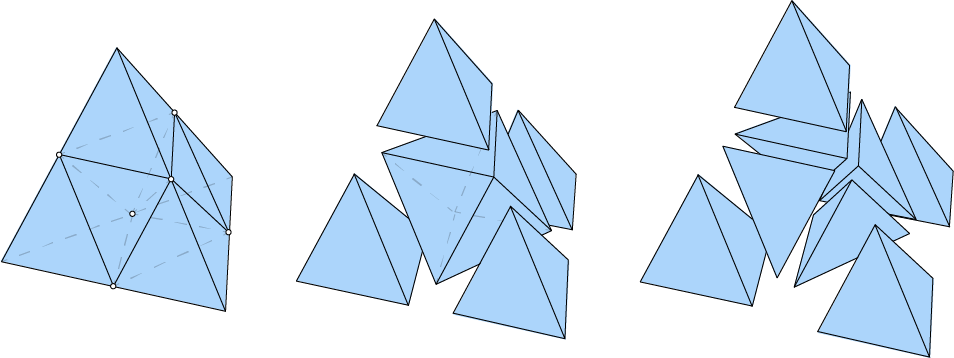}
    \caption{Subdivision structure.}
    \label{fig:subdiv}
    \end{minipage}
    \hfill
    \begin{minipage}[c]{0.48\columnwidth}
    \centering
    \includegraphics[width=0.8\columnwidth]{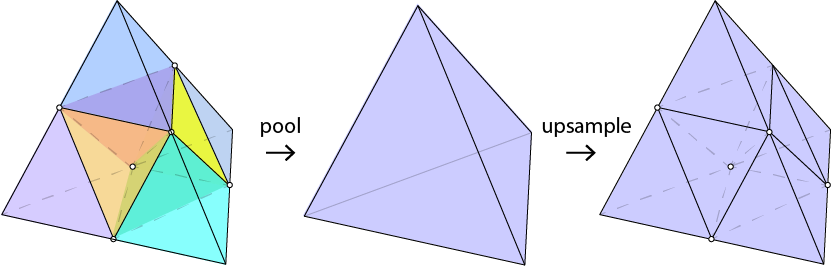}
    \caption{Pooling and upsampling operators.}
    \label{fig:pool_unpool}
    \end{minipage}
    \vspace{-0.6cm}
\end{figure}

To upsample features $\phi^{(n)}$ from $\tT{(n-1)}$ to $\tT^{(n)}$, one may define techniques analogous to those in 2D computer vision. %
We propose an alternative method analogous to interpolation upsampling. The upsampled features of $\mT_k^{(n)}\in\gS_j$ are an average of the features of the neighbors of its parent $\mT_j^{(n-1)}$, weighted by the inverse distance between centroids:
{\small
\bea
    \phi^{(n)}(\mT_k^{(n)}) = \sum_{\ell \in \gN(\mT_j^{(n-1)})} \frac{\alpha_1}{\norm{c_\ell^{(n-1)} - c_k^{(n)}}} \phi(T_\ell^{(n-1)}),
\eea
}
where $\alpha_1$ is a scalar that normalizes the coefficients to sum to one. For an illustration of the downsamping and upsampling processes, see~\figref{fig:pool_unpool} 

\subsection{TetGAN}\label{sec:tetgan}
{\bf Problem formulation.}
We represent the 3D shape by inserting it inside the irregular grid $\tT^{(N)}$ and computing an occupancy and a displacement per tetrahedron. The occupancy $\mO$ is given by:
$
    \mO(\mT^{(N)}_k) = \mathbf{1}[w_{\mM}\left(c_k^{(N)}\right) > 0],
$
where $w(\cdot)$ gives the winding number of any point with respect to the surface $\mM$ and $\mathbf{1}$ denotes the indicator function. Here, $c_k^{(N)}$ denotes the centroid of $\mT_k^{(N)}$ given by the average coordinate of its four vertices.
We define a deformation on a \textit{tetrahedron} $\mT_k^{(N)}$ as the difference between $c_k^{(N)}$ and its closest point $p$ on $\mM$. In order to obtain a deformation over the \textit{vertices}, we compute the average displacement over each tetrahedra incident to each vertex $v^{(N)}$, weighted by the inverse distance:
{\small
\bea
    \mD(v^{(N)}) = \sum_{k \in \gI(v^{(N)})} \frac{\alpha_2}{\norm{v^{(N)} - c_k^{(N)}}} \argmin_{p\in \mM} \norm{c_k^{(N)} - p} - c_k^{(N)},
\eea
}
where $\gI(v)$ denotes a set of incident tetrahedra indices of vertex $v$ and $\alpha_2$ again is a normalizing scalar. We directly penalize the deformation per-vertex against the ground-truth deformation field.
To represent a \textit{tetrahedral} mesh $\mM$ we have $\mX = [\mO, \mD]$. $\mO$ trivially defines a \textit{triangular mesh} consisting of all faces shared by tetrahedra with different occupancies. %
Our goal is to learn a generative model $p(\mX)$ from which $\mX$ can be sampled.

\textbf{Generative model.}
Our generative model is a hybrid between a VAE and a GAN. Starting with a VAE, we have an encoder $f_{\tt enc}$, a decoder $f_{\tt dec}$, and a prior function $p(\mZ)$.
Both encoder and decoder are deep neural networks consisting of the proposed tetrahedral convolution layers; exact architecture details are deferred to the supplemental. To generate a new mesh during inference, we can easily sample a latent vector $\mZ ~\sim p(\mZ)$ and pass it through the decoder $f_{\tt dec}(\mZ)$. We train the VAE using binary cross-entropy on the occupancies and mean squared loss on the deformations.
To enable a more densely sampled and smoothly varying latent space, we additionally include global and local (WGAN-GP~\cite{gulrajani2017improved}) adversarial losses. This introduces two discriminator networks $f_{\tt dis}^l$ and $f_{\tt dis}^g$, which operate over local and global attributes. The local discriminator is a fully convolutional \textit{patch discriminator}~\cite{li2016precomputed} which outputs a `real' or `fake' score per tetrahedron. On the other hand, the global discriminator makes use of pooling layers to aggregate information across the entire grid into a `real' or `fake' score.

\textbf{Deformation-Field Weighted Laplacian Smoothing.}
We apply Laplacian smoothing to the output mesh using weights computed with the predicted deformation field. Specifically, %
we first update the vertices by applying the deformation field over them, \ie $\hat{v}_i^{(N)} = v_i^{(N)} + D(v_i^{(N)})$. Next, these displaced vertices are smoothed by performing a weighted average with their neighbors:
{\small
\be
    \tilde{v}_i^{(N)} = \beta \cdot \hat{v}_i^{(N)} + (1-\beta)\cdot\sum_{\hat{v}_j^{(N)} \in \gN(\hat{v}_i^{(N)})} \left|\cos\left(\mD(v_i^{(N)}),\mD(v_j^{(N)})\right)\right| \cdot \hat{v}_j^{(N)}
\ee
}
where $\cos$ denotes the cosine similarity, $\gN(v_i^{(N)})$ 
denotes the neighboring vertices of $\hat{v}_i^{(N)}$ and $\beta \in [0, 1]$ is a hyperparameter that controls how much influence the original position of $\hat{v}_i^{(N)}$ exerts on its smoothed position $\tilde{v}_i^{(N)}$.

\textbf{Deformation field filtering.}
We observe that for tetrahedra that are truly on the surface of the underlying shape, the corresponding deformation should be relatively small in magnitude. Thus, we can use the predicted deformation field to filter incorrect occupancies. In particular, we compute the average norm of surface tetrahedra deformations in the dataset: 
\be
    \mu = \frac{1}{|D|} \sum_{\mM\in D} \frac{1}{|\mM|} \sum_{\substack{
    T_k \in \tT, \mO_\mM(T_k) = 1\\
    \exists T_j \in \gN_k\colon \mO_\mM(T_j) = 0
    }}\norm{\mD(T_k)}
\ee
where $|\mM|$ gives the number of occupied tetrahedra on the surface of $\mM$. During inference (but not during training), we simply discard surface tetrahedra with a deformation larger than $\gamma\mu$ where $\gamma$ is a tunable parameter.

\begin{figure}[t]
    \centering
    \begin{minipage}[c]{0.48\columnwidth}
    \includegraphics[width=\columnwidth, keepaspectratio,]{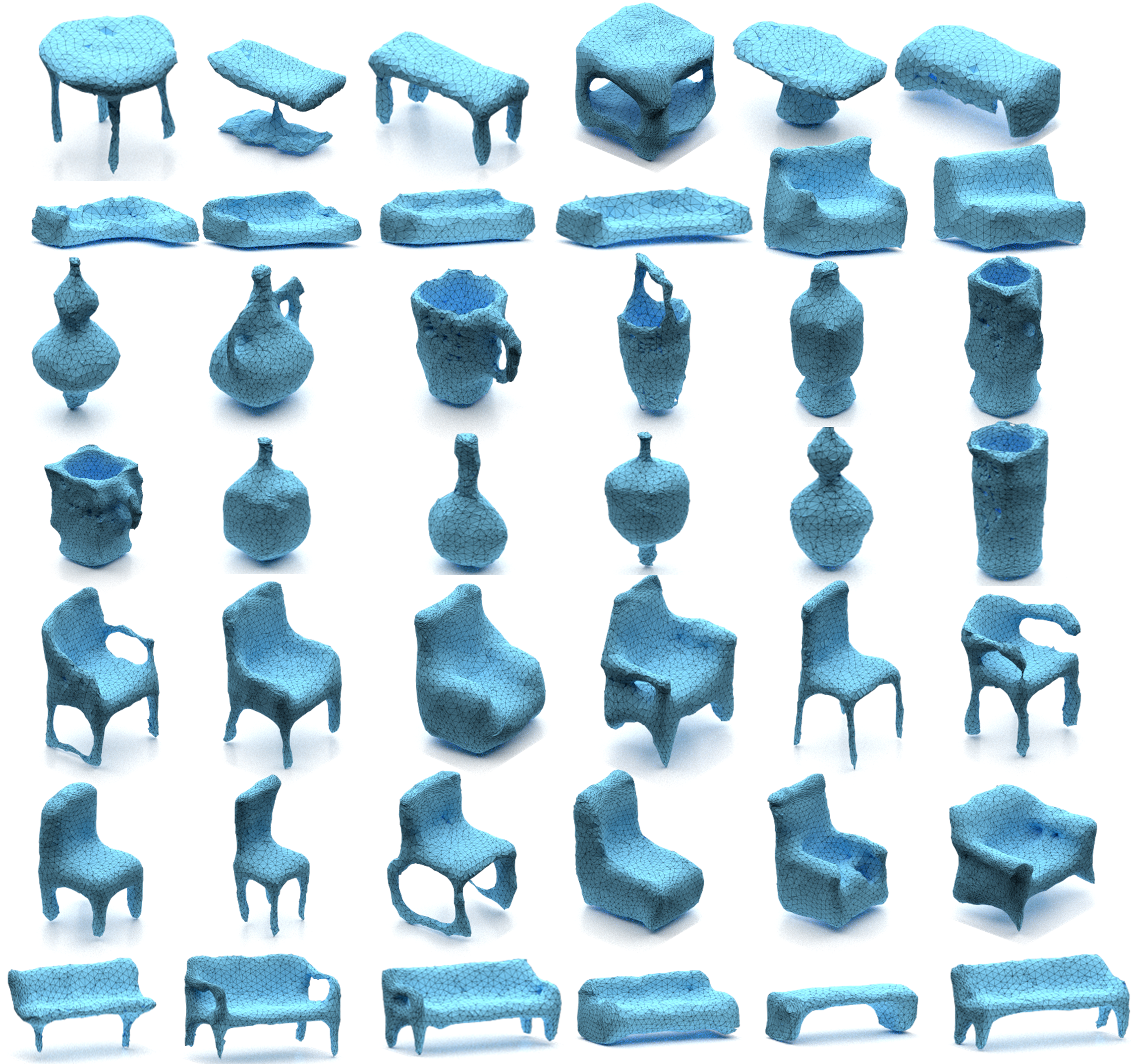}
    \small
    \caption{Novel random samples from TetGAN learned latent distribution.
    }
    \label{fig:sampled_chairs}
    \end{minipage}
    \hfill
    \begin{minipage}[c]{0.48\columnwidth}
    \centering
     \newcommand{\pl}{102}
    \newcommand{\plb}{-4}
    \begin{overpic}[width=\columnwidth]{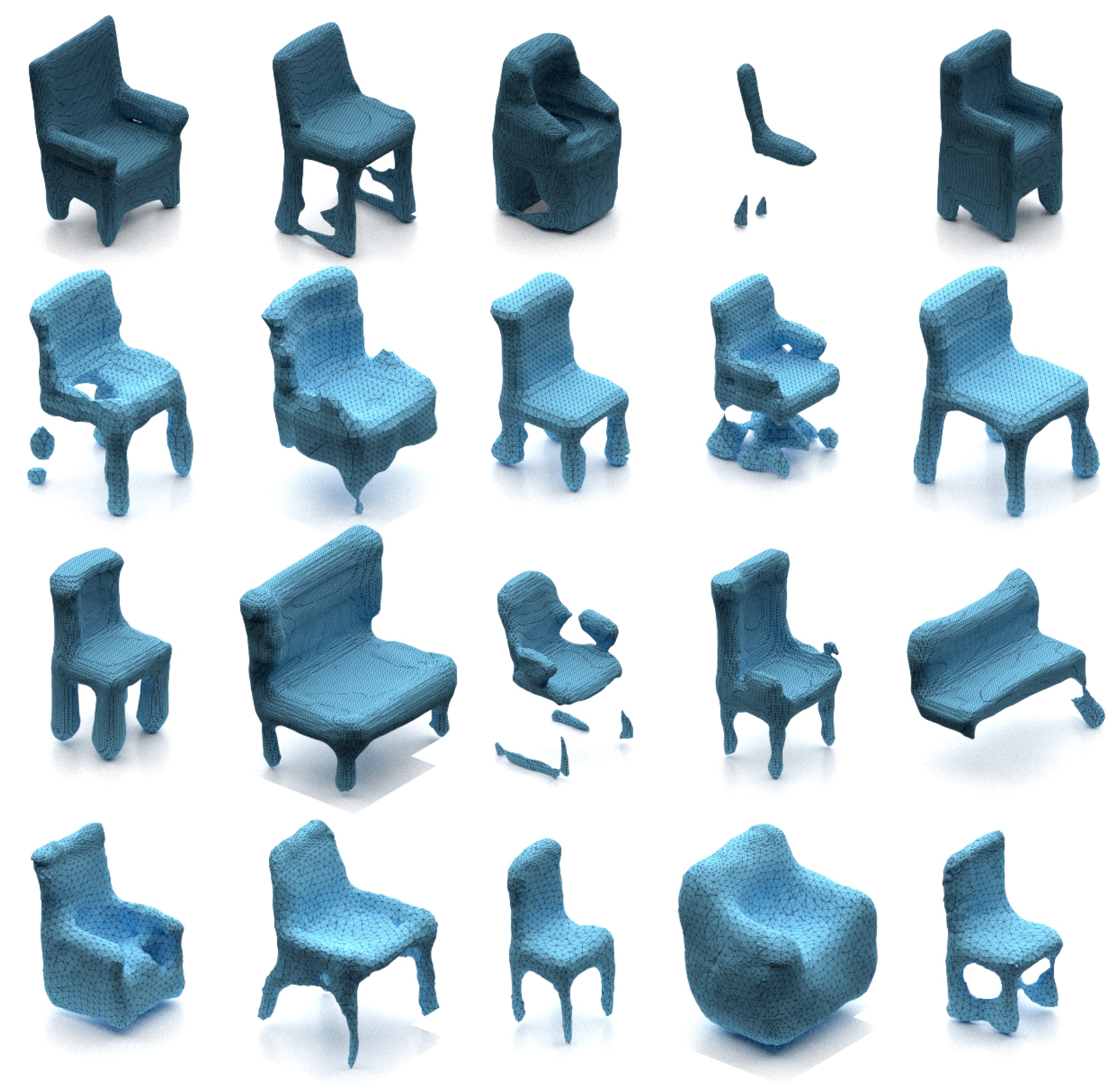}
    \put(\plb,  85){\textcolor{black}{(a)}}
    \put(\plb,  65){\textcolor{black}{(b)}}
    \put(\plb,  42){\textcolor{black}{(c)}}
    \put(\plb,  15){\textcolor{black}{(d)}}
    \end{overpic}
    \vspace{-0.6cm}
    \small
    \caption{Novel random samples from (a) OccNet, (b) ShapeGAN (VAE), and (c) ShapeGAN, and (d) \ourmethod{}.}
    \label{fig:samplescompare}
    \end{minipage}
    \vspace{-0.4cm}
\end{figure}
\section{Experiments}
\label{sec:experiments}
In our experiments, we assess the ability of the encoder to effectively map input shapes to the learned latent space, and the decoder to accurately reconstruct them. Next, we evaluate the variety of the generated shape samples when sampling from the learned latent distribution. We compare TetGAN against implicit networks~\cite{mescheder2019occupancy, kleineberg2020adversarial} quantitatively using Frechet Inception Distance (FID). Third, we qualitatively explore the latent space through latent space arithmetic and interpolation. 
Finally, we conduct ablation studies on our own system to quantify the impact of the proposed components,~\eg without adversarial losses. Note that there are no existing works that take tetrahedral meshes as input directly. Therefore, when comparing to baselines, we use their respective choice of discretization. 

\textbf{Dataset.} We evaluate \ourmethod{} on five ShapeNet~\cite{shapenet2015} categories (chair, sofa, benches, table, and car), and vases from COSEG~\cite{wang2012active}.

To prepare the training and validation data, we use a randomized $5/1$ train/validation split on the dataset.
We process the ShapeNet models to be manifold and watertight using the method by~\citet{huang2018robust}, and normalized to a unit bounding cube. We compute occupancy and deformation features for each shape as specified in~\secref{sec:tetgan}.

\subsection{Qualitative results}
\textbf{Visual Sample Quality.} To demonstrate the generative capabilities of TetGAN, in~\figref{fig:sampled_chairs} we show random samples over the categories of chair, sofa, table, benches, and vase. These samples are generated unconditionally (\ie purely from noise). 
TetGAN is trained separately for each category. We observe that \ourmethod{} is capable of generating a diverse set of shapes with varied topology. We also show qualitative comparison with the OccNet~\cite{mescheder2019occupancy} and ShapeGAN~\cite{kleineberg2020adversarial} in~\figref{fig:samplescompare}. %

\begin{figure*}[t]
    \centering
    \includegraphics[width=\textwidth]{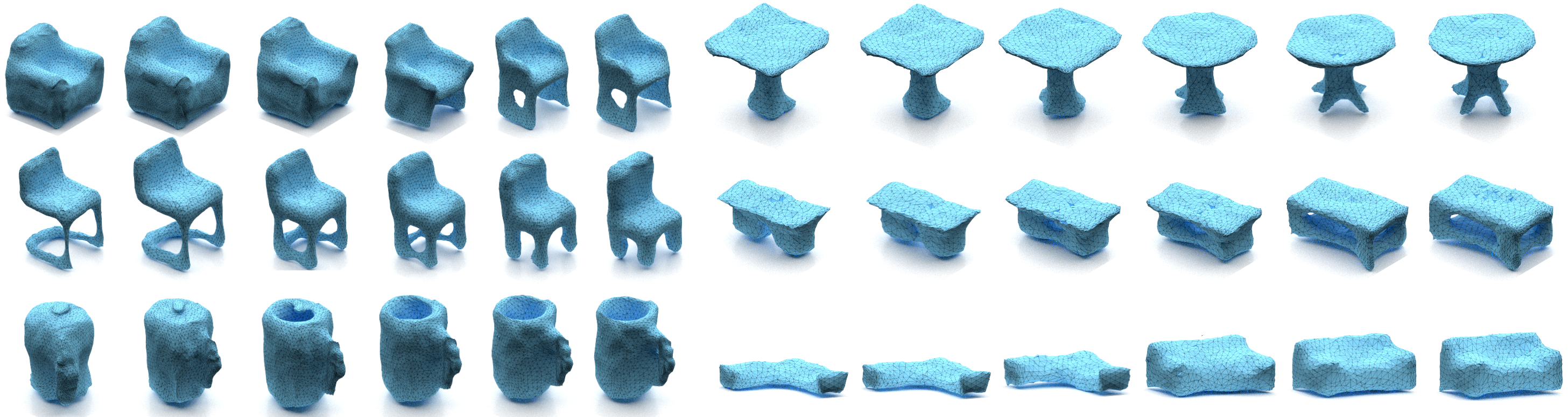}
    \vspace{-0.4cm}
    \caption{\ourmethod{}'s latent space interpolation %
    for different categories. 
    }
    \label{fig:interpolations}
    \vspace{-0.3cm}
\end{figure*}

\begin{wrapfigure}{R}{0.4\textwidth}
    \centering
    \vspace{-0.4cm}
     \newcommand{\acolor}{\color[rgb]{0.4,0.0,0.63}}
     \newcommand{\bcolor}{\color[rgb]{1,0.8,0.51}}
     \newcommand{\ccolor}{\color[rgb]{0.94,0.57,0.7}}
     \newcommand{\dcolor}{\color[rgb]{0.6,0.92,0.6}}
    \includegraphics[width=0.4\columnwidth]{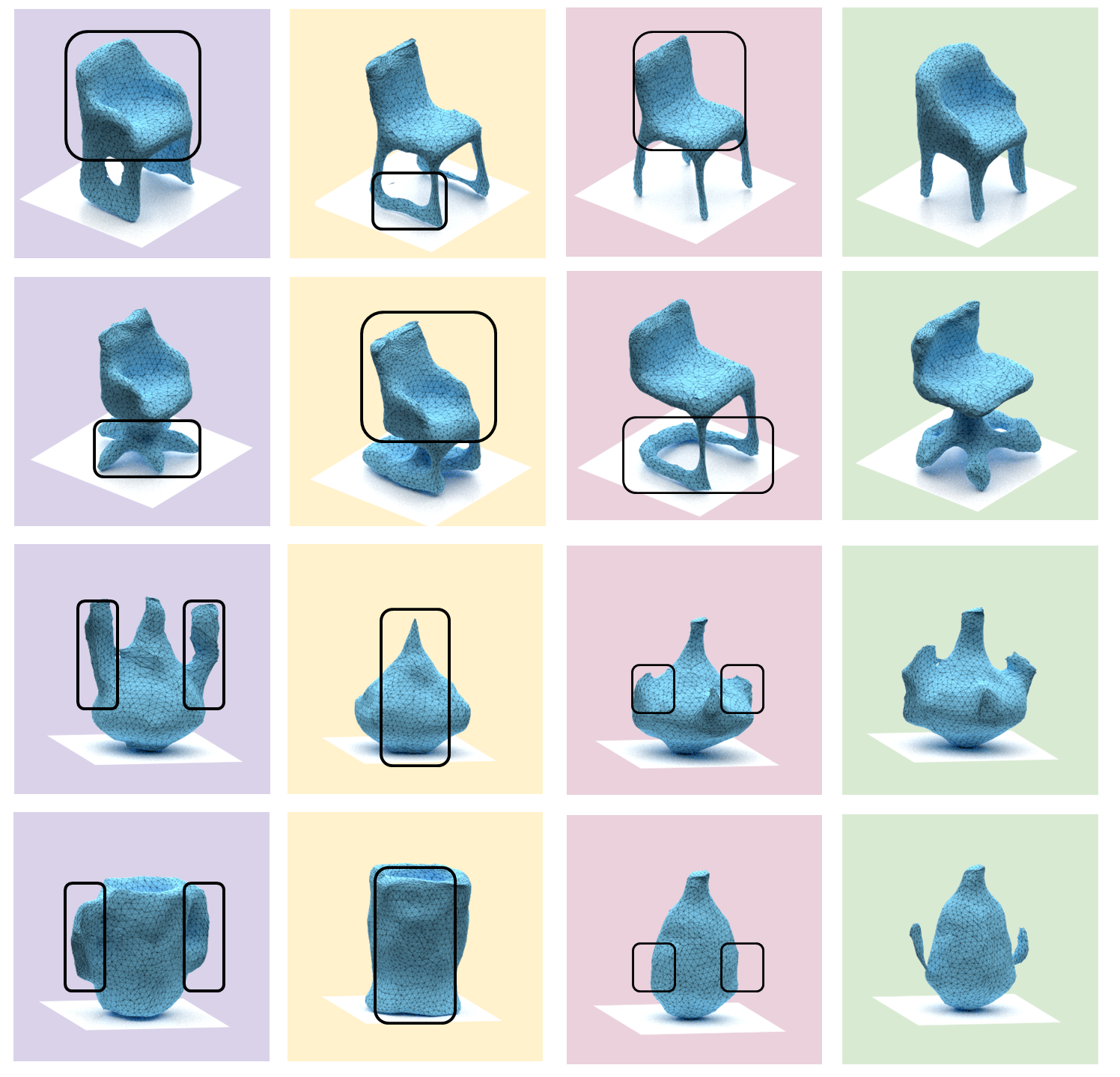}
    \vspace{-0.5cm}
    \caption{Latent space arithmetic between {\acolor \textbf{mesh A}} - {\bcolor \textbf{mesh B}} + {\ccolor \textbf{mesh C}} = {\dcolor \textbf{mesh D}}. The highlighted areas indicate: a part that is ({\acolor \textbf{added}}), {\bcolor \textbf{removed}}, and where it is {\ccolor \textbf{inserted}}. %
    }
    \label{fig:operations}
\end{wrapfigure}

\textbf{Latent Space Exploration.}
We experiment with the latent space~\ourmethod{} learns over shapes.
In~\figref{fig:interpolations} we demonstrate~\ourmethod{}'s ability to produce semantically meaningful intermediates between two shapes. 
We further demonstrate~\ourmethod{} encodes high-level semantics of objects and parts through latent space arithmetic.

As illustrated in~\figref{fig:operations}, the difference between two meshes can be added to another mesh resulting in a completely new object. For example, in the first row, by subtracting the encodings of two similar chairs (in purple and yellow backgrounds) with one main difference (in the chair legs), we are able to transfer that difference onto a third chair (in pink) resulting in a brand-new chair (in green).

\textbf{Utility of TetGAN Meshes.} We inspect the synthesized/interpolated meshes by slicing (\figref{fig:teaser} and the supplemental), and find that our network predicts meshes with a solid interior. Since these meshes also inherit a high tetrahedral quality from the underlying grid, they are useful for performing engineering simulations (see supplemental).

\subsection{Quantitative results}\label{sec:exp_gen}
~\tabref{tab:comparison} contains quantitative metrics evaluating reconstruction and generation of chairs. Note that all metrics evaluate \textit{surface quality} since baseline generative methods, unlike \ourmethod{}, do not produce volumetric meshes. See supplemental for additional quantitative evaluation.

\begin{wraptable}{R}{7.8cm}
    \vspace{-0.5cm}
    \setlength{\tabcolsep}{1pt}
    \small
    \centering    
    \begin{tabular}{@{}l ccccc@{}}
        \toprule
        \multirow{2}{*}{\textbf{Model}} &  \textbf{Recon.~} & \textbf{Avg.~} & \multirow{2}{*}{\textbf{FID}$\downarrow$} & \multirow{2}{*}{\textbf{Variety$\uparrow$}}  \\
        & \textbf{(Chamfer)}$\downarrow$ & \textbf{\# Faces}$\downarrow$ & &\\ %
        \midrule
        OccNet & $0.0015$ & $86901$ & \textbf{1.06} & \textbf{0.0041} \\
        ShapeGAN (GAN) & - & $25837$ & $3.38$ & $0.0036$ \\
        ShapeGAN (VAE) & \textbf{0.0011} & $6115$ & $8.66$ & $0.0017$ \\
        \hline
        OccNet* %
        & $0.019$ & $5500$ & $1.59$ & $0.0039$ \\
        ShapeGAN (GAN*) %
        & - & $5500$ & $3.87$ & $0.0038$ \\
        ShapeGAN (VAE*) %
        & \textbf{0.0012} & $5500$ & $9.25$ & $0.0016$ \\
        \ourmethod{} $-$ HR/wLS & 0.004 & $5263$ & $2.18$ & \textbf{0.0040} \\
        \ourmethod{} $-$ HR/No wLS & $0.0035$ & $5263$ &  \textbf{1.35} & $0.0028$ \\
        \ourmethod{} $-$ LR/wLS & $0.0032$ & \textbf{3062} &  $5.01$ & $0.0038$ \\
        \ourmethod{} $-$ LR/No wLS & $0.0029$ & \textbf{3062} &  $2.09$ & $0.0033$ \\
        \bottomrule
    \end{tabular}
    
    \caption{%
    HR/LR refer to grid resolution of $(61^3)$/$(41^3)$. wLS refers to our deformation-field weighted Laplacian smoothing.  * indicates mesh simplification of outputs. Note that these metrics are computed over surfaces since baseline methods do not generate volumes. Even so, \ourmethod{} outperforms alternatives at generation in comparable object resolutions (\ie number of faces).
    } 
    \label{tab:comparison}
    \vspace{-.6cm}
\end{wraptable}

\textbf{Reconstruction.}
We evaluate how accurately \ourmethod{} reconstructs shapes on a test set %
using Chamfer Distance as a metric for accuracy.~\tabref{tab:comparison} shows the Chamfer distance between the reconstructed and the ground truth objects as well as the number of faces in the reconstructed object of \ourmethod{} and baseline works. We observe that both OccNet and the VAE version of ShapeGAN beat \ourmethod{} in reconstruction.
However, for a fair comparison, we simplify OccNet generated meshes to a comparable number of faces as those produced by \ourmethod{}. In this setting, \ourmethod{} outperforms OccNet in reconstruction. We also note that, although \ourmethod{} does not beat ShapeGAN (VAE) this metric, reconstruction quality does not necessarily correlate with visual sample quality.

\textbf{Variety}. We evaluate the diversity of~\ourmethod{}'s shape generation by measuring the most similar pairs in a set of generated shapes. We generate $k$ pairs of shapes and take the average Chamfer distance from the $n$ most similar pairs. In our experiments, we use $k=250$ and $n=25$. We run the experiment $5$ times to account for variance. We observe that \ourmethod{} produces more varied shapes than both versions of ShapeGAN and comes close to the variety of OccNet, beating the simplified version.  

\textbf{Frechet Inception Distance.} We use \textbf{FID}~\cite{NIPS2017_8a1d6947} between a set of generated shapes and the hold-out set to measure quality of generated shapes. To compute this distance, we use deep features from a PointNet++~\cite{qi2017pointnet} architecture trained for classification. We observe that at a high grid resolution, \ourmethod{} without Laplacian smoothing comes close to the FID that OccNet achieves and beats both versions of ShapeGAN. \ourmethod{} again beats the simplfied version of OccNet. \ourmethod{} at low grid resolution also produces shapes of reasonable quality and FID score. We observe that Laplacian smoothing, although produces desirable results, harms the FID score. This is likely due to the smoothing producing unnaturally thin pieces.

\subsection{Ablation Study}\label{sec:ablation}
\begin{figure}[t]
    \begin{minipage}[c]{0.48\columnwidth}
    \vspace{0.25cm}
    \centering
     \newcommand{\pl}{85}
    \newcommand{\plb}{-6}
    \begin{overpic}[width=0.95\columnwidth]{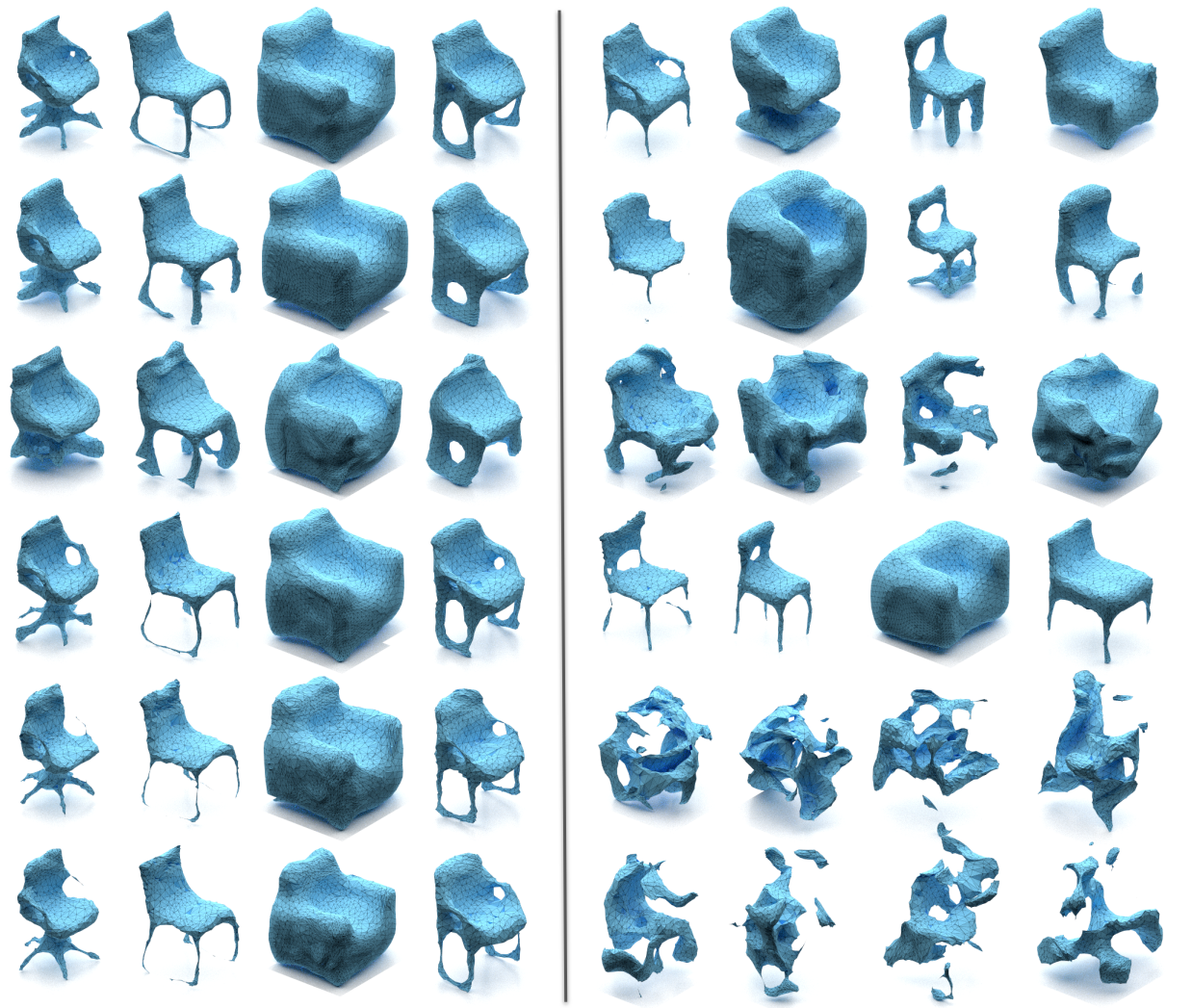}
    \put(6,  \pl){\textcolor{black}{Reconstruction}}
    \put(65,  \pl){\textcolor{black}{Samples}}
    \put(\plb,  78){\textcolor{black}{(a)}}
    \put(\plb,  64){\textcolor{black}{(b)}}
    \put(\plb,  50){\textcolor{black}{(c)}}
    \put(\plb,  35){\textcolor{black}{(d)}}
    \put(\plb,  20){\textcolor{black}{(e)}}
    \put(\plb,  5){\textcolor{black}{(f)}}
    \end{overpic}
    \vspace{-0.1cm}
    \caption{Ablations. Top to bottom: (a) full \ourmethod{}, (b) surface GCN, (c) tetrahedral GCN, (d) no patch discriminator, (e) no global discriminator, (f) VAE only.%
    }
    \label{fig:ablations}
    \end{minipage}
    \hspace{3pt}
    \begin{minipage}[c]{0.48\columnwidth}
    \vspace{-0.1cm}
    \centering
    \newcommand{\pl}{-3}
    \newcommand{\plb}{-6}
    \newcommand{\plbb}{-9}
    \begin{overpic}[width=\columnwidth]{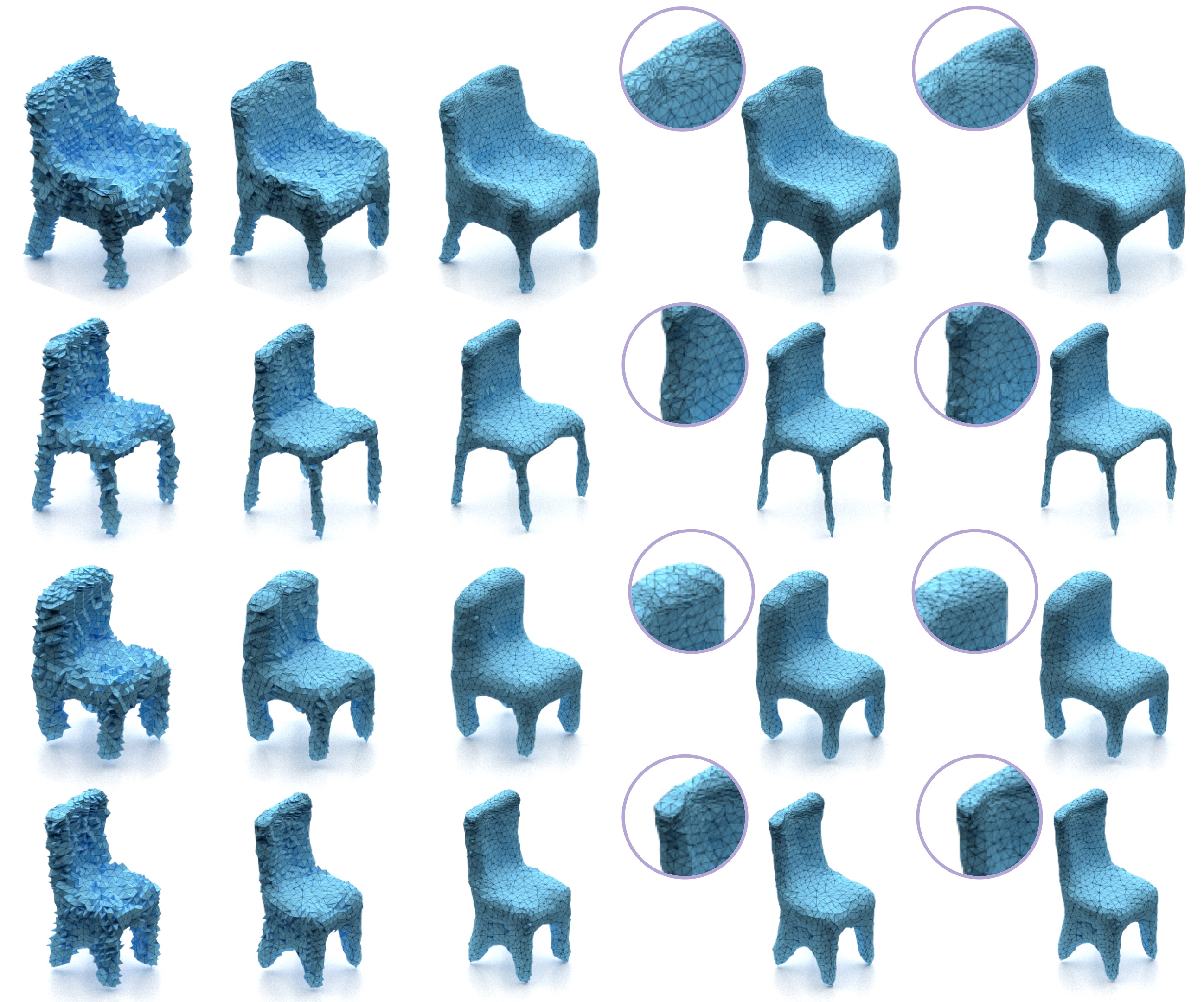}
    \put(0,  -3){\textcolor{black}{No Def.}}
    \put(21,  -3){\textcolor{black}{Def.}}
    \put(35,  -3){\textcolor{black}{No Def.}}
    \put(39,  -8){\textcolor{black}{Our}}
    \put(35, -12){\textcolor{black}{Smooth}}
    \put(64,  -3){\textcolor{black}{Def.}}
    \put(61.5,  -8){\textcolor{black}{Cotan}}
    \put(60, -12){\textcolor{black}{Smooth}}
    \put(85,  -3){\textcolor{black}{Def.}}
    \put(85,  -8){\textcolor{black}{Our}}
    \put(80,  -12){\textcolor{black}{Smooth}}
    \end{overpic}
    \vspace{0.3ex}
    \caption{Our deformation-field weighted smooth (Def. Our Smooth) improves the quality of the final result. 
    }
    \label{fig:deformation_smooth}
    \end{minipage}
\end{figure}

\textbf{Adversarial ablations.}
We evaluate the effectiveness of the components in \ourmethod{}. 
\ourmethod{} performs best using the full system, although the global discriminator appears far more important to the quality of generated results. Visually, as seen in \figref{fig:ablations}, removing the local discriminator (row $4$) causes floating artifacts. Removing the global discriminator (row $5$) ruins the generative ability of \ourmethod{}. We find that the global discriminator is crucial to learning the global structure of shapes. This is supported by a sharp increase in FID ($23.35$ without the global discriminator, only $12.27$ without the local discriminator).

\textbf{Graph convolutions.} In~\figref{fig:ablations}, row (b), we train using the GCN used in DefTet~\cite{deftet}. Their method trains a GCN to predict displacements over the vertices of a surface extracted from the ground-truth meshes. In this setting, we observe that the predicted deformations tend to be quite small and not well-suited for a noisy occupancy field. In~\figref{fig:ablations} row (c), we learn a tetrahedral GCN (learning the same weight for each the four neighbors), and observe significant quality degradation.

\textbf{Deformation and smoothing.}
We experimented with different methods of smoothing the output of \ourmethod{}. As seen in \figref{fig:deformation_smooth}, the learned displacements were often were not sufficient to create a smooth mesh. However, combining the vertex displacements with an additional weighted Laplacian smoothing (wLS) procedure yields desirable results, smoother and better preservation of detail compared to LS alone. Comparing the last two columns, we also observe that our wLS creates sharper features compared to cotangent LS.

\section{Conclusion}
\label{sec:conclusion}
We presented a technique for generating tetrahedral meshes, which learns to encode occupancy and displacement field over an irregular grid of tetrahedra. This formulation enables building neural networks capable of tetrahedral generation and manipulation. 
Our approach combines advantages from both adversarial learning and variational encoding, enabling learning semantically meaningful latent spaces which can synthesize novel and diverse shapes with varied topology using a  small number of faces. \ourmethod{} utilizes a unique spatial structure for 3D shapes to build layers which contain a powerful \textit{inductive bias}. 

A current limitation of our system is that volumetric grids have a large memory footprint, especially compared to images. An interesting direction for future work involves incorporating a hierarchical subdivision scheme, which would enable selectively adding resolution where needed. In the future we are interested in exploring a technique for generative tetrahedral meshes using an alternative tetrahedral meshing strategy, such as TetWild~\cite{tetwild} or TetGen~\cite{TetGen} to learn to predict tetrahedra whose size is adaptive to the volume.

\textbf{Acknowledgements}
This work was partially supported by gifts from Adobe Research. This work was supported through the University of Chicago AI cluster resources, services, and staff expertise. We thank Greg Shakhnarovich, Vova Kim, Daniele Panozzo, Haochen Wang, Meitar Shechter, Kfir Aberman, and the \emph{3DL group} for their helpful comments, suggestions, and insightful discussions. We thank Guanyue Rao and Xiaosheng Gao for their help in performing physical simulations. We thank Zihan (Jack) Zhang for assistance running experiments, and the anonymous reviewers for their comments.

\newpage
\bibliography{bibs}

\begin{thebibliography}{54}
\providecommand{\natexlab}[1]{#1}
\providecommand{\url}[1]{\texttt{#1}}
\expandafter\ifx\csname urlstyle\endcsname\relax
  \providecommand{\doi}[1]{doi: #1}\else
  \providecommand{\doi}{doi: \begingroup \urlstyle{rm}\Url}\fi

\bibitem[Bridson and Doran(2013)]{quartet}
Robert Bridson and Crawford Doran.
\newblock Quartet.
\newblock \url{https://github.com/crawforddoran/quartet}, 2013.

\bibitem[Chang et~al.(2015)Chang, Funkhouser, Guibas, Hanrahan, Huang, Li,
  Savarese, Savva, Song, Su, Xiao, Yi, and Yu]{shapenet2015}
Angel~X. Chang, Thomas Funkhouser, Leonidas Guibas, Pat Hanrahan, Qixing Huang,
  Zimo Li, Silvio Savarese, Manolis Savva, Shuran Song, Hao Su, Jianxiong Xiao,
  Li~Yi, and Fisher Yu.
\newblock {{ShapeNet}: An Information-Rich 3D Model Repository}.
\newblock Technical Report arXiv:1512.03012 [cs.GR], Stanford University ---
  Princeton University --- Toyota Technological Institute at Chicago, 2015.

\bibitem[Chen and Zhang(2019)]{chen2019learning}
Zhiqin Chen and Hao Zhang.
\newblock Learning implicit fields for generative shape modeling.
\newblock In \emph{Proc. IEEE Conf. Computer Vision and Pattern Recognition},
  2019.

\bibitem[Chen et~al.(2020)Chen, Tagliasacchi, and Zhang]{chen2020bsp}
Zhiqin Chen, Andrea Tagliasacchi, and Hao Zhang.
\newblock Bsp-net: Generating compact meshes via binary space partitioning.
\newblock In \emph{Proc. IEEE Conf. Computer Vision and Pattern Recognition},
  2020.

\bibitem[Chen et~al.(2021)Chen, Kim, Fisher, Aigerman, Zhang, and
  Chaudhuri]{chen2021decor}
Zhiqin Chen, Vladimir~G Kim, Matthew Fisher, Noam Aigerman, Hao Zhang, and
  Siddhartha Chaudhuri.
\newblock {DECOR-GAN}: {3D} shape detailization by conditional refinement.
\newblock In \emph{Proc. IEEE Conf. Computer Vision and Pattern Recognition},
  2021.

\bibitem[Cutler et~al.(2002)Cutler, Dorsey, McMillan, M{\"u}ller, and
  Jagnow]{cutler2002procedural}
Barbara Cutler, Julie Dorsey, Leonard McMillan, Matthias M{\"u}ller, and Robert
  Jagnow.
\newblock A procedural approach to authoring solid models.
\newblock \emph{ACM Transactions on Graphics}, 2002.

\bibitem[Esser et~al.(2021)Esser, Rombach, and Ommer]{vqgan}
Patrick Esser, Robin Rombach, and Bjorn Ommer.
\newblock Taming transformers for high-resolution image synthesis.
\newblock In \emph{Proc. IEEE Conf. Computer Vision and Pattern Recognition},
  2021.

\bibitem[Gao et~al.(2020)Gao, Chen, Xiang, Jacobson, McGuire, Fidler, Ling,
  Acuna, Kreis, Kim, et~al.]{deftet}
Jun Gao, Wenzheng Chen, Tommy Xiang, Alec Jacobson, Morgan McGuire, Sanja
  Fidler, Huan Ling, David Acuna, Karsten Kreis, Seung Kim, et~al.
\newblock Learning deformable tetrahedral meshes for {3D} reconstruction.
\newblock In \emph{Advances in Neural Info. Proc. Systems}, 2020.

\bibitem[Gao et~al.(2019)Gao, Yang, Wu, Yuan, Fu, Lai, and Zhang]{gao2019sdm}
Lin Gao, Jie Yang, Tong Wu, Yu-Jie Yuan, Hongbo Fu, Yu-Kun Lai, and Hao Zhang.
\newblock Sdm-net: Deep generative network for structured deformable mesh.
\newblock \emph{ACM Transactions on Graphics}, 2019.

\bibitem[Goodfellow et~al.(2014)Goodfellow, Pouget-Abadie, Mirza, Xu,
  Warde-Farley, Ozair, Courville, and Bengio]{goodfellow2014generative}
Ian Goodfellow, Jean Pouget-Abadie, Mehdi Mirza, Bing Xu, David Warde-Farley,
  Sherjil Ozair, Aaron Courville, and Yoshua Bengio.
\newblock Generative adversarial nets.
\newblock In \emph{Advances in Neural Info. Proc. Systems}, 2014.

\bibitem[Groueix et~al.(2018)Groueix, Fisher, Kim, Russell, and
  Aubry]{groueix2018papier}
Thibault Groueix, Matthew Fisher, Vladimir~G Kim, Bryan~C Russell, and Mathieu
  Aubry.
\newblock A papier-m{\^a}ch{\'e} approach to learning {3D} surface generation.
\newblock In \emph{Proc. IEEE Conf. Computer Vision and Pattern Recognition},
  2018.

\bibitem[Gulrajani et~al.(2017)Gulrajani, Ahmed, Arjovsky, Dumoulin, and
  Courville]{gulrajani2017improved}
Ishaan Gulrajani, Faruk Ahmed, Martin Arjovsky, Vincent Dumoulin, and Aaron
  Courville.
\newblock Improved training of {Wasserstein} {GANs}.
\newblock In \emph{Advances in Neural Info. Proc. Systems}, 2017.

\bibitem[Hanocka et~al.(2019)Hanocka, Hertz, Fish, Giryes, Fleishman, and
  Cohen-Or]{hanocka2019meshcnn}
Rana Hanocka, Amir Hertz, Noa Fish, Raja Giryes, Shachar Fleishman, and Daniel
  Cohen-Or.
\newblock {MeshCNN}: a network with an edge.
\newblock \emph{ACM Transactions on Graphics}, 2019.

\bibitem[Hanocka et~al.(2020)Hanocka, Metzer, Giryes, and
  Cohen-Or]{Hanocka2020p2m}
Rana Hanocka, Gal Metzer, Raja Giryes, and Daniel Cohen-Or.
\newblock {Point2Mesh}: A self-prior for deformable meshes.
\newblock \emph{ACM Transactions on Graphics}, 2020.

\bibitem[Hertz et~al.(2020)Hertz, Hanocka, Giryes, and Cohen-Or]{hertz2020deep}
Amir Hertz, Rana Hanocka, Raja Giryes, and Daniel Cohen-Or.
\newblock Deep geometric texture synthesis.
\newblock \emph{ACM Transactions on Graphics}, 2020.

\bibitem[Heusel et~al.(2017)Heusel, Ramsauer, Unterthiner, Nessler, and
  Hochreiter]{NIPS2017_8a1d6947}
Martin Heusel, Hubert Ramsauer, Thomas Unterthiner, Bernhard Nessler, and Sepp
  Hochreiter.
\newblock {GAN}s trained by a two time-scale update rule converge to a local
  nash equilibrium.
\newblock In \emph{Advances in Neural Info. Proc. Systems}, 2017.

\bibitem[Hu et~al.(2021)Hu, Liu, Guo, Cai, Huang, Mu, and Martin]{subdivnet}
Shi-Min Hu, Zheng-Ning Liu, Meng-Hao Guo, Jun-Xiong Cai, Jiahui Huang,
  Tai-Jiang Mu, and Ralph~R. Martin.
\newblock Subdivision-based mesh convolution networks.
\newblock \emph{ACM Transactions on Graphics}, 2021.

\bibitem[Hu et~al.(2018)Hu, Zhou, Gao, Jacobson, Zorin, and Panozzo]{tetwild}
Yixin Hu, Qingnan Zhou, Xifeng Gao, Alec Jacobson, Denis Zorin, and Daniele
  Panozzo.
\newblock Tetrahedral meshing in the wild.
\newblock \emph{ACM Transactions on Graphics}, 2018.

\bibitem[Hu et~al.(2020)Hu, Schneider, Wang, Zorin, and Panozzo]{ftetwild}
Yixin Hu, Teseo Schneider, Bolun Wang, Denis Zorin, and Daniele Panozzo.
\newblock Fast tetrahedral meshing in the wild.
\newblock \emph{ACM Transactions on Graphics}, 2020.

\bibitem[Huang et~al.(2018)Huang, Su, and Guibas]{huang2018robust}
Jingwei Huang, Hao Su, and Leonidas Guibas.
\newblock Robust watertight manifold surface generation method for {ShapeNet}
  models.
\newblock \emph{arXiv preprint arXiv:1802.01698}, 2018.

\bibitem[Jacobson et~al.(2011)Jacobson, Baran, Popovic, and
  Sorkine]{jacobson2011bounded}
Alec Jacobson, Ilya Baran, Jovan Popovic, and Olga Sorkine.
\newblock Bounded biharmonic weights for real-time deformation.
\newblock \emph{ACM Transactions on Graphics}, 2011.

\bibitem[Jatavallabhula et~al.(2021)Jatavallabhula, Macklin, Golemo, Voleti,
  Petrini, Weiss, Considine, Parent-Levesque, Xie, Erleben,
  et~al.]{jatavallabhula2021gradsim}
Krishna~Murthy Jatavallabhula, Miles Macklin, Florian Golemo, Vikram Voleti,
  Linda Petrini, Martin Weiss, Breandan Considine, Jerome Parent-Levesque,
  Kevin Xie, Kenny Erleben, et~al.
\newblock {gradSim}: Differentiable simulation for system identification and
  visuomotor control.
\newblock In \emph{Intl. Conf. on Learning Representations}, 2021.

\bibitem[Karras et~al.(2019)Karras, Laine, and Aila]{stylegan}
Tero Karras, Samuli Laine, and Timo Aila.
\newblock A style-based generator architecture for generative adversarial
  networks.
\newblock In \emph{Proc. IEEE Conf. Computer Vision and Pattern Recognition},
  2019.

\bibitem[Karras et~al.(2021)Karras, Aittala, Laine, H\"ark\"onen, Hellsten,
  Lehtinen, and Aila]{stylegan3}
Tero Karras, Miika Aittala, Samuli Laine, Erik H\"ark\"onen, Janne Hellsten,
  Jaakko Lehtinen, and Timo Aila.
\newblock Alias-free generative adversarial networks.
\newblock In \emph{Advances in Neural Info. Proc. Systems}, 2021.

\bibitem[Kingma and Ba(2015)]{KingmaB14}
Diederik~P. Kingma and Jimmy Ba.
\newblock Adam: A method for stochastic optimization.
\newblock In \emph{Intl. Conf. on Learning Representations}, 2015.

\bibitem[Kingma and Welling(2014)]{vae}
Diederik~P. Kingma and Max Welling.
\newblock {Auto-Encoding Variational Bayes}.
\newblock In \emph{International Conference on Learning Representations}, 2014.

\bibitem[Kleineberg et~al.(2020)Kleineberg, Fey, and
  Weichert]{kleineberg2020adversarial}
Marian Kleineberg, Matthias Fey, and Frank Weichert.
\newblock Adversarial generation of continuous implicit shape representations.
\newblock In \emph{Conf. of the European Association for Computer Graphics},
  2020.

\bibitem[Lahav and Tal(2020)]{lahav2020meshwalker}
Alon Lahav and Ayellet Tal.
\newblock Meshwalker: Deep mesh understanding by random walks.
\newblock \emph{ACM Transactions on Graphics}, 2020.

\bibitem[Li and Wand(2016)]{li2016precomputed}
Chuan Li and Michael Wand.
\newblock Precomputed real-time texture synthesis with {Markovian} generative
  adversarial networks.
\newblock In \emph{Proc. European Conf. on Computer Vision}, 2016.

\bibitem[Liu et~al.(2020)Liu, Kim, Chaudhuri, Aigerman, and
  Jacobson]{liu2020neural}
Hsueh-Ti~Derek Liu, Vladimir~G Kim, Siddhartha Chaudhuri, Noam Aigerman, and
  Alec Jacobson.
\newblock Neural subdivision.
\newblock \emph{ACM Transactions on Graphics}, 2020.

\bibitem[Mehralian and Karasfi(2018)]{DCGAN}
Mehran Mehralian and Babak Karasfi.
\newblock Rdcgan: Unsupervised representation learning with regularized deep
  convolutional generative adversarial networks.
\newblock In \emph{AIAR}, 2018.

\bibitem[Mescheder et~al.(2019)Mescheder, Oechsle, Niemeyer, Nowozin, and
  Geiger]{mescheder2019occupancy}
Lars Mescheder, Michael Oechsle, Michael Niemeyer, Sebastian Nowozin, and
  Andreas Geiger.
\newblock Occupancy networks: Learning {3D} reconstruction in function space.
\newblock In \emph{Proc. IEEE Conf. Computer Vision and Pattern Recognition},
  2019.

\bibitem[Milano et~al.(2020)Milano, Loquercio, Rosinol, Scaramuzza, and
  Carlone]{pdmeshnet}
Francesco Milano, Antonio Loquercio, Antoni Rosinol, Davide Scaramuzza, and
  Luca Carlone.
\newblock Primal-dual mesh convolutional neural networks.
\newblock In \emph{Advances in Neural Info. Proc. Systems}, 2020.

\bibitem[Munkberg et~al.(2022)Munkberg, Hasselgren, Shen, Gao, Chen, Evans,
  Mueller, and Fidler]{munkberg2021nvdiffrec}
Jacob Munkberg, Jon Hasselgren, Tianchang Shen, Jun Gao, Wenzheng Chen, Alex
  Evans, Thomas Mueller, and Sanja Fidler.
\newblock {Extracting Triangular {3D} Models, Materials, and Lighting From
  Images}.
\newblock In \emph{Proc. IEEE Conf. Computer Vision and Pattern Recognition},
  2022.

\bibitem[Nash et~al.(2020)Nash, Ganin, Eslami, and Battaglia]{nash2020polygen}
Charlie Nash, Yaroslav Ganin, SM~Ali Eslami, and Peter Battaglia.
\newblock Polygen: An autoregressive generative model of {3D} meshes.
\newblock In \emph{Intl. Conf. on Machine Learning}, 2020.

\bibitem[Paill\'e et~al.(2015)Paill\'e, Ray, Poulin, Sheffer, and
  L\'evy]{PRPSL15}
Gilles-Philippe Paill\'e, Nicolas Ray, Pierre Poulin, Alla Sheffer, and Bruno
  L\'evy.
\newblock Dihedral angle-based maps of tetrahedral meshes.
\newblock \emph{ACM Transactions on Graphics}, 2015.

\bibitem[Park et~al.(2019)Park, Florence, Straub, Newcombe, and
  Lovegrove]{park2019deepsdf}
Jeong~Joon Park, Peter Florence, Julian Straub, Richard Newcombe, and Steven
  Lovegrove.
\newblock Deep{SDF}: Learning continuous signed distance functions for shape
  representation.
\newblock In \emph{Proc. IEEE Conf. Computer Vision and Pattern Recognition},
  2019.

\bibitem[Paschalidou et~al.(2021)Paschalidou, Katharopoulos, Geiger, and
  Fidler]{paschalidou2021neural}
Despoina Paschalidou, Angelos Katharopoulos, Andreas Geiger, and Sanja Fidler.
\newblock Neural parts: Learning expressive {3D} shape abstractions with
  invertible neural networks.
\newblock In \emph{Proc. IEEE Conf. Computer Vision and Pattern Recognition},
  2021.

\bibitem[Peng et~al.(2021)Peng, Jiang, Liao, Niemeyer, Pollefeys, and
  Geiger]{sap}
Songyou Peng, Chiyu~"Max" Jiang, Yiyi Liao, Michael Niemeyer, Marc Pollefeys,
  and Andreas Geiger.
\newblock Shape as points: A differentiable {Poisson} solver.
\newblock In \emph{Advances in Neural Info. Proc. Systems}, 2021.

\bibitem[Qi et~al.(2017)Qi, Su, Mo, and Guibas]{qi2017pointnet}
Charles~R Qi, Hao Su, Kaichun Mo, and Leonidas~J Guibas.
\newblock Pointnet: Deep learning on point sets for {3D} classification and
  segmentation.
\newblock In \emph{Proc. IEEE Conf. Computer Vision and Pattern Recognition},
  2017.

\bibitem[Rakotosaona et~al.(2021)Rakotosaona, Aigerman, Mitra, Ovsjanikov, and
  Guerrero]{rakotosaona2021differentiable}
Marie-Julie Rakotosaona, Noam Aigerman, Niloy~J Mitra, Maks Ovsjanikov, and
  Paul Guerrero.
\newblock Differentiable surface triangulation.
\newblock \emph{ACM Transactions on Graphics}, 2021.

\bibitem[Sawhney and Crane(2020)]{sawhney2020mcgp}
Rohan Sawhney and Keenan Crane.
\newblock Monte carlo geometry processing: A grid-free approach to {PDE}-based
  methods on volumetric domains.
\newblock \emph{ACM Transactions on Graphics}, 2020.

\bibitem[Sharp and Ovsjanikov(2020)]{sharp2020pointtrinet}
Nicholas Sharp and Maks Ovsjanikov.
\newblock {PointTriNet}: Learned triangulation of 3{D} point sets.
\newblock In \emph{Proc. European Conf. on Computer Vision}, 2020.

\bibitem[Shen et~al.(2021)Shen, Gao, Yin, Liu, and Fidler]{shen2021dmtet}
Tianchang Shen, Jun Gao, Kangxue Yin, Ming-Yu Liu, and Sanja Fidler.
\newblock Deep marching tetrahedra: a hybrid representation for high-resolution
  {3D} shape synthesis.
\newblock In \emph{Advances in Neural Info. Proc. Systems}, 2021.

\bibitem[Si(2015)]{TetGen}
Hang Si.
\newblock {TetGen}, a {Delaunay}-based quality tetrahedral mesh generator.
\newblock \emph{ACM Trans. Math. Softw.}, 2015.

\bibitem[Trusty et~al.(2021)Trusty, Chen, and
  Levin]{Trusty_Chen_Levin_SEM_2021}
Ty~Trusty, Honglin Chen, and David~I.W. Levin.
\newblock The shape matching element method: Direct animation of curved surface
  models.
\newblock \emph{ACM Transactions on Graphics}, 2021.

\bibitem[Tulsiani et~al.(2017)Tulsiani, Su, Guibas, Efros, and
  Malik]{tulsiani2017learning}
Shubham Tulsiani, Hao Su, Leonidas~J Guibas, Alexei~A Efros, and Jitendra
  Malik.
\newblock Learning shape abstractions by assembling volumetric primitives.
\newblock In \emph{Proc. IEEE Conf. Computer Vision and Pattern Recognition},
  2017.

\bibitem[Uesu et~al.(2005)Uesu, Bavoil, Fleishman, Shepherd, and
  Silva]{uesu2005simplification}
Dirce Uesu, Louis Bavoil, Shachar Fleishman, Jason Shepherd, and Cl{\'a}udio~T
  Silva.
\newblock Simplification of unstructured tetrahedral meshes by point sampling.
\newblock In \emph{Intl. Workshop on Volume Graphics}, 2005.

\bibitem[Wang et~al.(2018)Wang, Zhang, Li, Fu, Liu, and Jiang]{pixel2mesh}
Nanyang Wang, Yinda Zhang, Zhuwen Li, Yanwei Fu, Wei Liu, and Yu-Gang Jiang.
\newblock {Pixel2Mesh}: Generating {3D} mesh models from single {RGB} images.
\newblock In \emph{Proc. European Conf. on Computer Vision}, 2018.

\bibitem[Wang et~al.(2015)Wang, Jacobson, Barbi{\v{c}}, and
  Kavan]{wang2015linear}
Yu~Wang, Alec Jacobson, Jernej Barbi{\v{c}}, and Ladislav Kavan.
\newblock Linear subspace design for real-time shape deformation.
\newblock \emph{ACM Transactions on Graphics}, 2015.

\bibitem[Wang et~al.(2012)Wang, Asafi, Van~Kaick, Zhang, Cohen-Or, and
  Chen]{wang2012active}
Yunhai Wang, Shmulik Asafi, Oliver Van~Kaick, Hao Zhang, Daniel Cohen-Or, and
  Baoquan Chen.
\newblock Active co-analysis of a set of shapes.
\newblock \emph{ACM Transactions on Graphics}, 2012.

\bibitem[Wu et~al.(2016)Wu, Zhang, Xue, Freeman, and Tenenbaum]{3dgan}
Jiajun Wu, Chengkai Zhang, Tianfan Xue, William~T Freeman, and Joshua~B
  Tenenbaum.
\newblock Learning a probabilistic latent space of object shapes via 3{D}
  generative-adversarial modeling.
\newblock In \emph{Advances in Neural Info. Proc. Systems}, 2016.

\bibitem[Yin et~al.(2020)Yin, Chen, Chaudhuri, Fisher, Kim, and
  Zhang]{yin2020coalesce}
Kangxue Yin, Zhiqin Chen, Siddhartha Chaudhuri, Matthew Fisher, Vladimir~G Kim,
  and Hao Zhang.
\newblock {COALESCE}: Component assembly by learning to synthesize connections.
\newblock In \emph{Intl. Conf. on 3D Vision}, 2020.

\bibitem[Zhu et~al.(2017)Zhu, Park, Isola, and Efros]{cycleGAN}
Jun-Yan Zhu, Taesung Park, Phillip Isola, and Alexei~A Efros.
\newblock Unpaired image-to-image translation using cycle-consistent
  adversarial networks.
\newblock In \emph{Proc. IEEE Conf. Computer Vision and Pattern Recognition},
  2017.

\end{thebibliography}
\newpage
\appendix
\section{Architecture details}\label{sec:architecture}
The network architecture of \ourmethod{} combines a convolutional variational autoencoder and two convolutional Wasserstein discriminators, which are denoted as $f_{\tt enc}$, $f_{\tt dec}$, $f_{\tt dis}^l$, and $f_{\tt dis}^g$.

The encoder $f_{\tt enc}$ consists of $N$ convolutional blocks of $4$ convolution layers, divided by $N-1$ pooling layers that downsample the irregular grid. After the final convolution, \ourmethod{} can be adapted to use a variety of modules to compute an encoding. In the variational setting, we also use $2$ additional linear layers, for the output layers of $f_{\tt enc}^\mu$ and $f_{\tt enc}^{\sigma^2}$. All layers except the final output layers are followed by instance normalization and a leaky ReLU nonlinearity.

The decoder $f_{\tt dec}$ is nearly a reversal of the encoder. It consists of $N$ convolutional blocks of 4 convolution layers, divided by $N-1$ upsampling layers that increase the grid resolution. Before the first convolution, we reverse the module that produces the final encoding. Again, all layers except the final output convolutional layer are followed by instance normalization and a leaky ReLU nonlinearity.

\ourmethod{} is fairly robust to the choice in module used in computing the final encoding. In this manuscript, we chose to use a large linear layer across the entire grid for the high-resolution models (making one latent code per mesh) and a shared MLP per-tetrahedron for the low-resolution models (making one latent code \textit{per tetrahedron}). In our experiments, using a Tetrahedral convolution to compute the final encoding also works well, although it produces thinner meshes. We choose a latent code size of $512$.  

The local discriminator $f^l_{\tt dis}$ is a PatchGAN~\cite{li2016precomputed} discriminator outputting a probability per tetrahedron $\mT_i^{(N)}\in \tT^{(N)}$. A final probability is computed by averaging over the tetrahedra. Specifically, $f^l_{\tt dis}$ has a fully convolutional architecture, where the final layer has $1$ output channel, representing the WGAN score per tetrahedron. Each convolutional layer except the final is followed by instance normalization and a leaky ReLU nonlinearity. 

The global discriminator $f^g_{\tt dis}$ is a more standard convolutional discriminator outputting a single probability for the entire grid $\tT^{(N)}$. Its architecture mirrors the encoder architecture, consisting of $N$ convolutional blocks of $3$ convolution layers, divided by $N-1$ pooling layers. After the final convolution, we apply a fully connected linear layer that outputs a scalar WGAN score. As before, each layer, except the final linear layer, is followed by instance normalization and a leaky ReLU nonlinearity.

\section{Implementation details}
\textbf{Subdivision.}
Note that there is no canonical way to subdivide a tetrahedron. The diagonal in the subdivision grid is selected based on the initial ordering of the vertices/edges in the low resolution grid (and remains fixed/unchanged throughout).

\textbf{Hyperparameters.} 
We subdivide the base tetrahedral grid $3$ times, i.e. choosing $N=4$. We trained~\ourmethod{} using the Adam optimizer~\cite{KingmaB14} with learning rate of $0.0001$ and $\beta_1, \beta_2 = 0, 0.9$ as suggested by \citet{gulrajani2017improved}. We choose a batch size of $30$. We choose $0.5$ for the weighted smoothing hyperparameter $\beta$ (main text, equation 4) and $4$ for the deformation field filtering hyperparameter $\gamma$ (main text, equation 5). 

\textbf{Runtime.} Each shape file takes on average 1-2 seconds to preprocess, depending on the size of the file. The total training time is approximately two days on two NVIDIA A40 GPUs for the larger categories (such as chairs). Smaller categories (such as benches) will give reasonable results in one day, and may not benefit from training any longer.

\section{Experiments}
\subsection{Dataset details}
\textbf{Dataset size.} The size of the dataset in COSEG is only 300 vases. For the ShapeNet categories, there are 6778 chairs, 1813 benches, 3173 sofas, 3514 cars, and 8436 tables.

\textbf{Non-manifold edges.}
Our method may contain non-manifold edges. We ran some statistics on our data and found that $98.9\%$ of edges in the ground-truth (chair) dataset are manifold, and $98.5\%$ of edges in the generated tetrahedral meshes are manifold. 

\subsection{Additional quantitative results}
We present additional quantitative results evaluating \ourmethod{} in unconditional generation using ShapeNet benches, sofas, tables, and COSEG vases. We train high-resolution (approx. $61^3$) models for benches and vases, and low-resolution (approx. $41^3$) models for sofas and tables, choosing resolution based on perceived necessity. We compare against unsimplified versions of the baselines.

\begin{wraptable}{r}{0.5\textwidth}
    \vspace{-0.5cm}
    \setlength{\tabcolsep}{2pt}
    \small
    \centering    
    \begin{tabular}{@{}l ccccc@{}}
        \toprule
        \textbf{Model} & \textbf{Benches} & \textbf{Sofas} & \textbf{Tables} & \textbf{Vases} \\
        \midrule
        OccNet & 10.91 & \textbf{1.54} & 2.47 & 7.42 \\
        ShapeGAN (GAN) & 7.46 & 1.57 & \textbf{0.48} & 5.46 \\
        ShapeGAN (VAE) & 7.74 & 8.30 & 6.48 & 20.51 \\
        \ourmethod{} & \textbf{2.67} & 3.35 & 2.13 & \textbf{2.57} \\
        \bottomrule
    \end{tabular}
    \caption{FID$\downarrow$ scores of \ourmethod{} and baselines on various categories. \ourmethod{} is best in 2 categories and competitive in the rest despite generating volumes (at a smaller object resolution) while baselines only generate surfaces.}
    \label{tab:supp_fid}
\end{wraptable}

\textbf{FID} Recall that we use we use deep features from a PointNet++~\cite{qi2017pointnet} architecture trained for classification to calculate FID score. \tabref{tab:supp_fid} shows the scores for \ourmethod{} and baselines across multiple categories. We observe that \ourmethod{} beats baselines in both the Shapenet benches and COSEG vases categories, models which are trained at a high grid resolution. \ourmethod{} is able to remain competitive with the \textit{surface-based} baselines, while task of generating \textit{coarse volumetric} meshes \ie tetrahedral meshes with fewer surface faces.

\begin{wraptable}{r}{0.5\textwidth}
    \vspace{-1cm}
    \setlength{\tabcolsep}{1pt}
    \small
    \centering    
    \begin{tabular}{@{}l ccccc@{}}
        \toprule
        \textbf{Model} & \textbf{Benches} & \textbf{Sofas} & \textbf{Tables} & \textbf{Vases} \\
        \midrule
        OccNet & 0.0014 & \textbf{0.0017} & \textbf{0.0053} & 0.0021 \\
        ShapeGAN (GAN) & \textbf{0.0017} & 0.0013 & 0.0047 & 0.0022 \\
        ShapeGAN (VAE) & 0.0003 & 0.0004 & 0.0032 & 0.0009 \\
        \ourmethod{} & 0.0014 & \textbf{0.0017} & 0.0045 & \textbf{0.0023} \\
        \bottomrule
    \end{tabular}
    \caption{Variety metric $\uparrow$ for \ourmethod{} and baselines on various categories. }
    \label{tab:supp_var}
\end{wraptable}

\textbf{Variety} Recall we evaluate variety by averaging the Chamfer distance between the $k$ closest pairs of generated meshes within a sample of size $N$ (choosing $k=25$, $N=250$ in practice). \tabref{tab:supp_var} shows the variety metrics for \ourmethod{} and baselines. We observe that \ourmethod{} again beats baselines in 2 categories, sofas and vases and is relatively competitive in the other two. 

\begin{wraptable}{r}{0.5\textwidth}
    \vspace{-1cm}
    \setlength{\tabcolsep}{1pt}
    \small
    \centering    
    \begin{tabular}{@{}l ccccc@{}}
        \toprule
        \textbf{Model} & \textbf{Benches} & \textbf{Sofas} & \textbf{Tables} & \textbf{Vases} \\
        \midrule
        ShapeGAN (VAE) & \textbf{0.0010} & \textbf{0.0018} & \textbf{0.0014} & \textbf{0.0048} \\
        \hline
        OccNet & 0.0095 & 0.0118 & 0.0222 & 0.0187 \\
        \ourmethod{} & \textbf{0.0046} & \textbf{0.0024} & \textbf{0.0029} & \textbf{0.0055} \\
        \bottomrule
    \end{tabular}
    \caption{Reconstruction accuracy $\downarrow$ for \ourmethod{} and baselines. \ourmethod{} and OccNet both have effective generative capabilities while still enabling reconstruction/encoding.}
    \label{tab:supp_recon}
\end{wraptable}

\textbf{Reconstruction} \tabref{tab:supp_recon} shows the average Chamfer distance between reconstructed meshes and ground truth meshes for \ourmethod{} and baselines over multiple categories. The VAE version of ShapeGAN, which is a pure VAE, beats the alternatives handily. However, \ourmethod{} offers a significant improvement over OccNet in all categories.

\FloatBarrier

\subsection{Additional visual results}
\begin{figure*}[h]
    \centering
    \includegraphics[width=\textwidth]{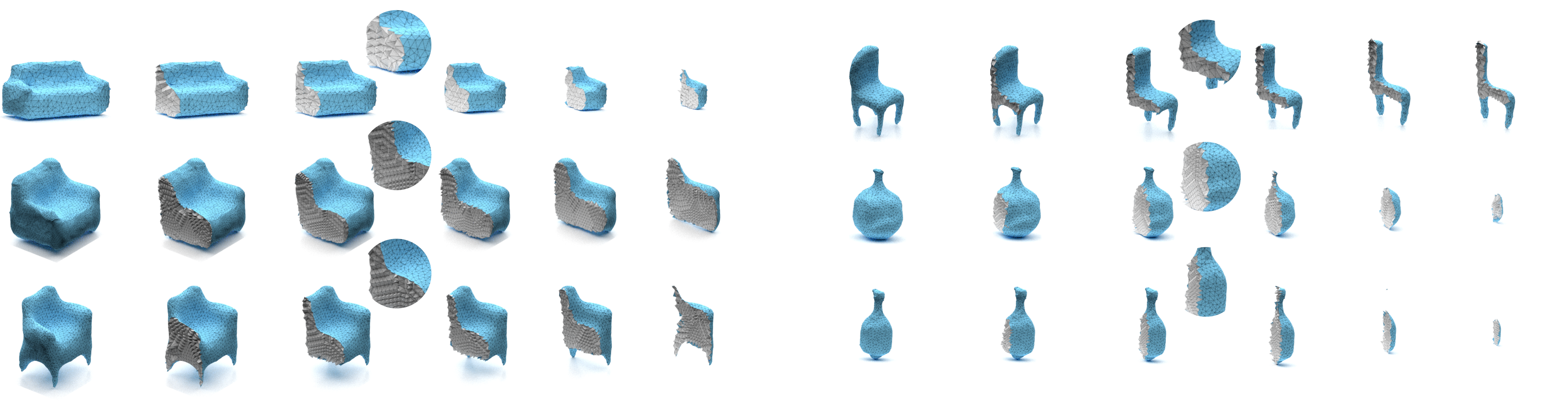}
    \caption{\ourmethod{} generates meshes with solid interiors.}
    \label{fig:tetdiscr0}
\end{figure*}

\begin{figure*}[h]
    \centering
    \includegraphics[width=\textwidth]{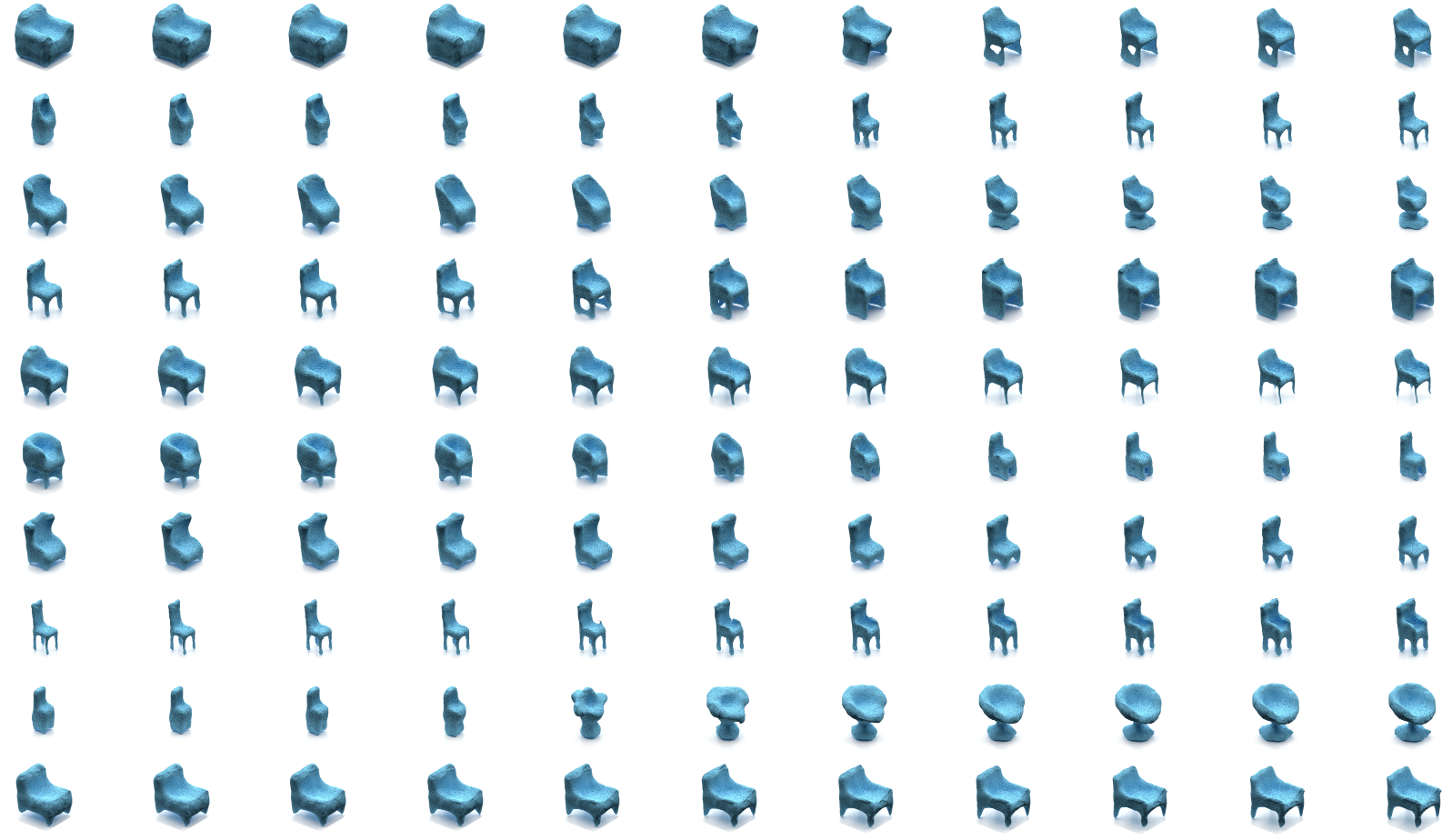}
    \caption{Additional chair interpolations.}
    \label{fig:tetdiscr1}
\end{figure*}

\begin{figure*}[h]
    \centering
    \includegraphics[width=\textwidth]{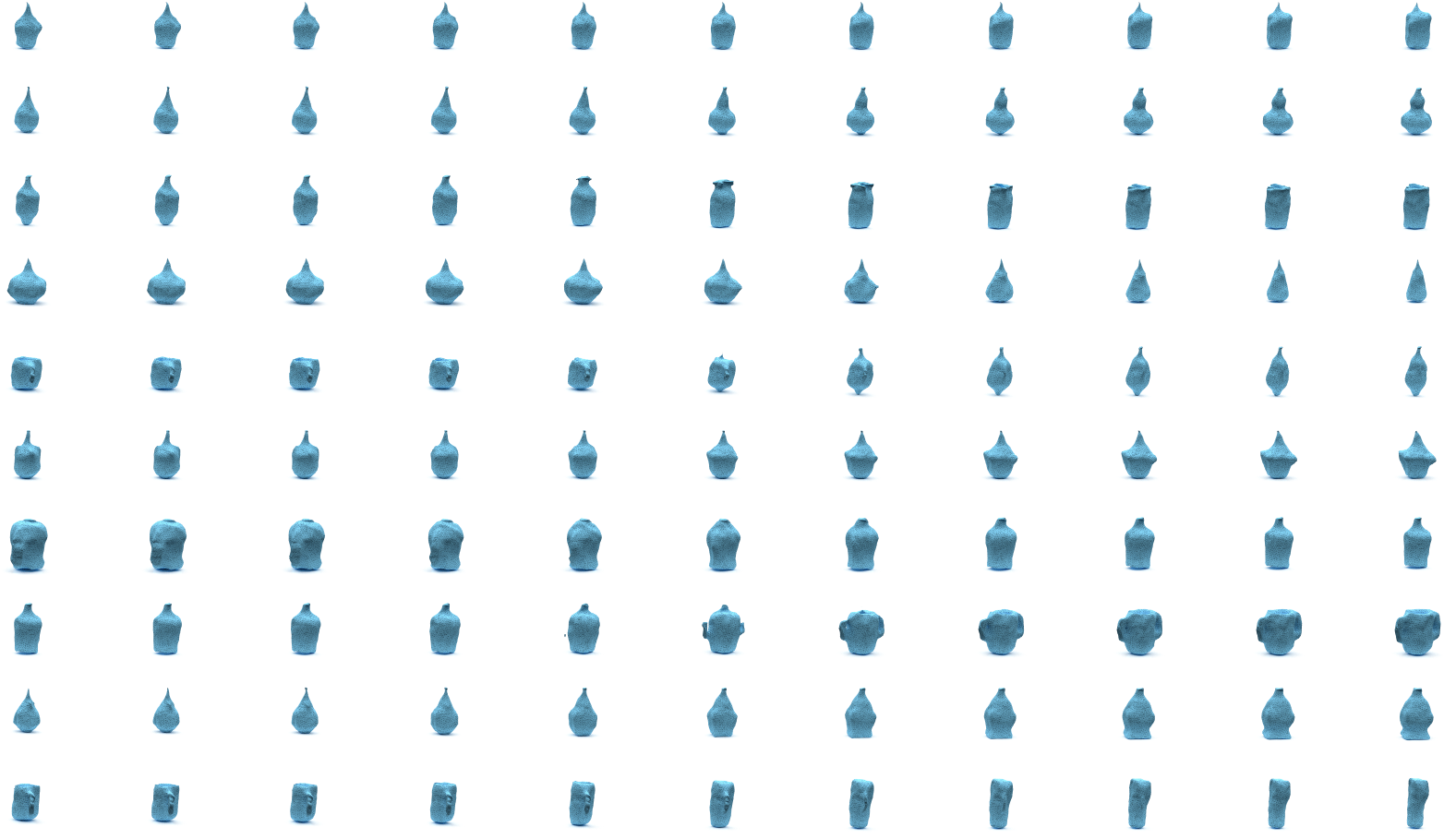}
    \caption{Additional vase interpolations.}
    \label{fig:tetdiscr2}
\end{figure*}

\begin{figure*}[h]
    \centering
\includegraphics[width=\columnwidth]{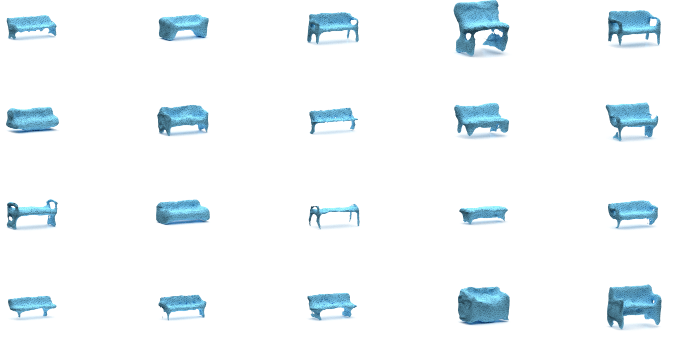}
    \caption{Unconditionally sampled benches.}
    \label{fig:tetdiscr3}
\end{figure*}

\begin{figure*}[h]
    \centering
\includegraphics[width=\columnwidth]{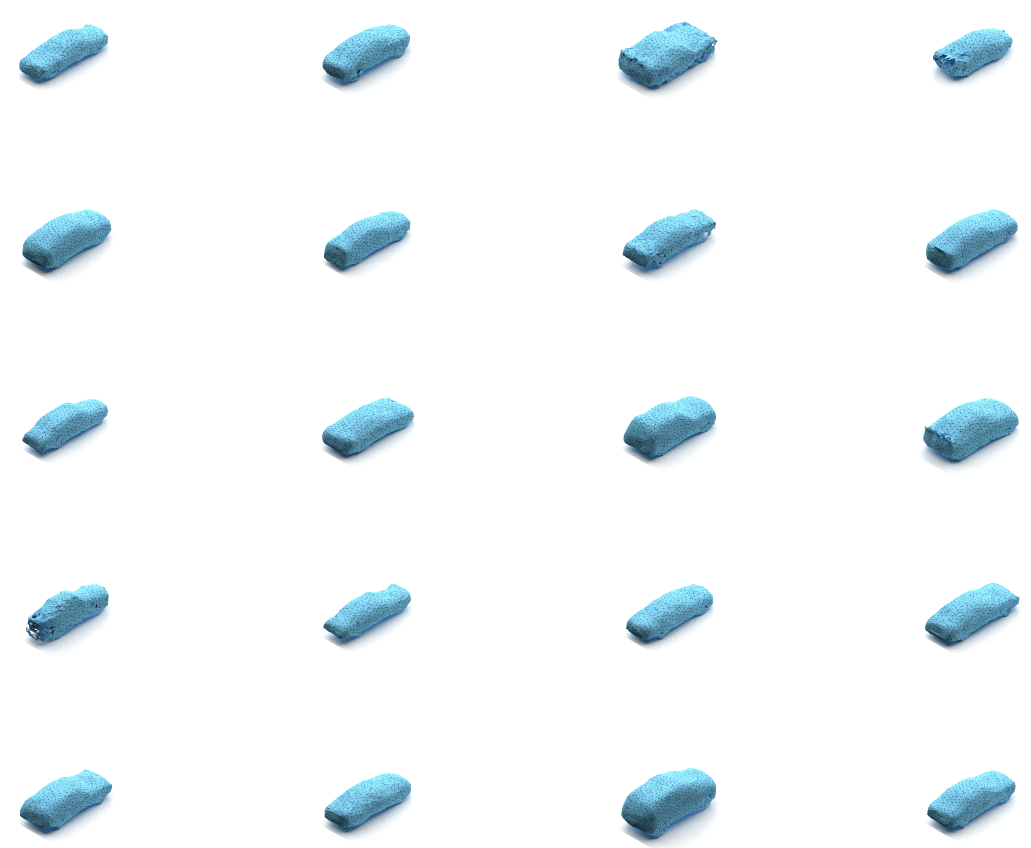}
    \caption{Unconditionally sampled cars. We observe that although the model is able to learn the coarse shape of a car, the discretization is not fine enough to capture details.}
    \label{fig:tetdiscr4}
\end{figure*}

\begin{figure}[t]
    \centering
    \vspace{-0.3cm}
    \includegraphics[width=\columnwidth]{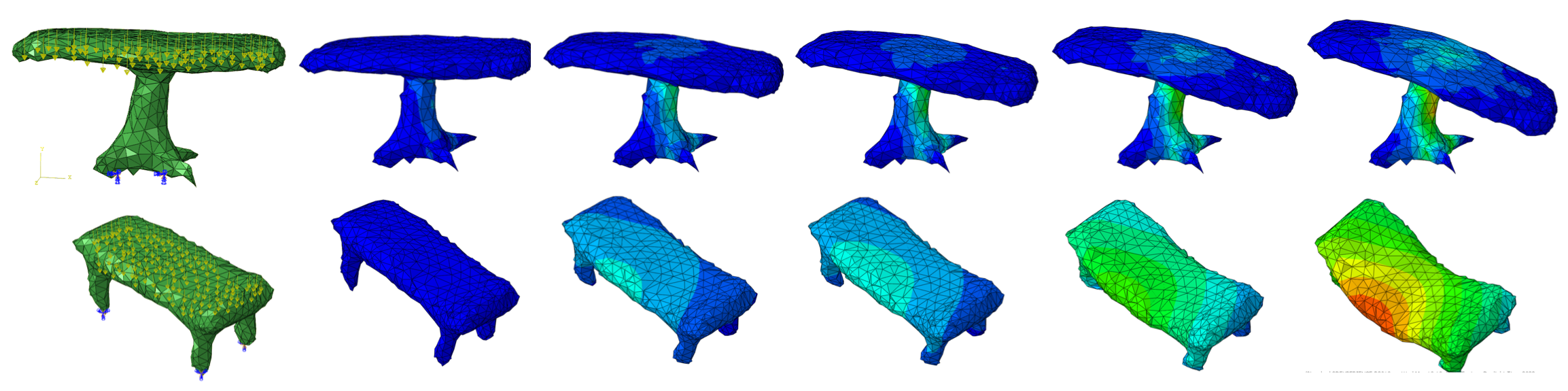}
    \vspace{-0.6cm}
    \caption{Simulation using TetGAN generated tetrahedral meshes. We apply a load to simulate a force acting on our generated table made out of pinewood and amplified the deformation by $2000\times$ for visualization.}
    \label{fig:simulation}
    \vspace{-0.4cm}
\end{figure}

\begin{figure}[h]
    \vspace{-0.5cm}
    \centering
    \includegraphics[width=\columnwidth]{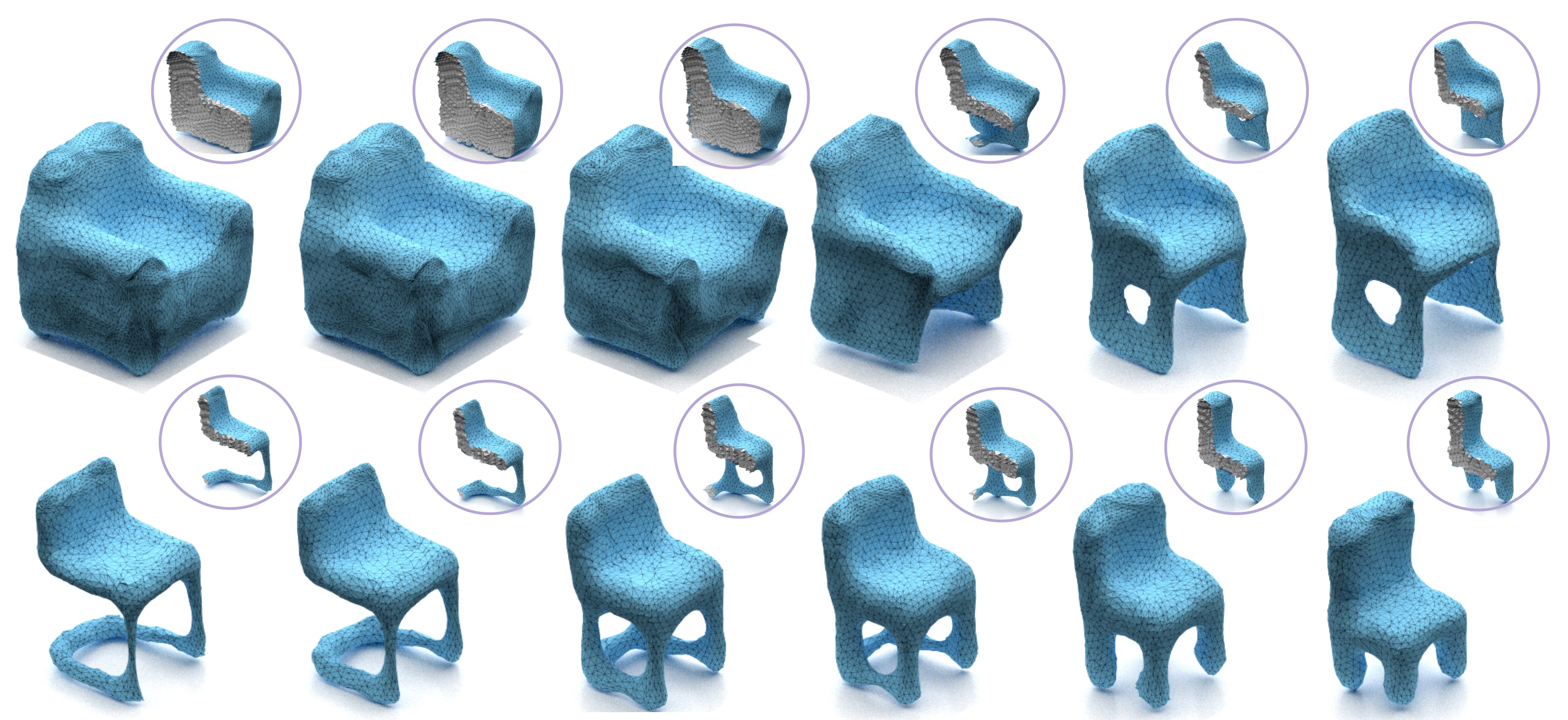}
    \caption{\ourmethod{} generates a series of smooth interpolations between two shapes with solid interiors. Inset: corresponding cross-section. \ourmethod{} produces solid interiors for novel interpolated shapes.}
    \label{fig:latent}
    \vspace{-0.5cm}
\end{figure}

\end{document}